\documentclass{article}

\PassOptionsToPackage{numbers, compress}{natbib}

\usepackage[preprint]{neurips_2026}

\usepackage[utf8]{inputenc} 
\usepackage[T1]{fontenc}    
\usepackage{hyperref}       
\usepackage{url}            
\usepackage{booktabs}       
\usepackage{amsfonts}       
\usepackage{nicefrac}       
\usepackage{microtype}      
\usepackage{xcolor}         
\usepackage{amsmath}
\usepackage{enumitem}
\setcitestyle{semicolon}
\usepackage{graphicx}
\usepackage{multirow}
\usepackage{titletoc}

\title{Navigating the Safety-Fidelity Trade-off: Massive-Variate Time
Series Forecasting for Power Systems via Probabilistic Scenarios}

\author{%
  Kaijie Xu \\
  ZJU-UIUC Institute\\
  Zhejiang University\\
  Hangzhou, China \\
  \texttt{Kaijie.23@intl.zju.edu.cn} \\
  \And
  Anqi Wang \\
  ZJU-UIUC Institute\\
  Zhejiang University\\
  Hangzhou, China \\
  \texttt{anqi.23@intl.zju.edu.cn} \\
  \And
  Xilin Dai\thanks{Corresponding author.} \\
  ZJU-UIUC Institute\\
  Zhejiang University\\
  Hangzhou, China \\
  \texttt{xilin2023@zju.edu.cn} \\
}

\begin{document}

\maketitle

\begin{abstract}

Probabilistic forecasting models are increasingly deployed on multivariate systems with distinct channel physics and operational constraints, but existing benchmarks evaluate neither property at scale. Public canonical multivariate benchmarks cap out at $2{,}000$ channels, while power-system benchmarks either lack temporal structure or probabilistic evaluation. We introduce PowerPhase, a probabilistic forecasting benchmark built on six transmission grids ranging from $2{,}000$ to $36{,}964$ jointly forecasted channels, more than an order of magnitude beyond popular canonical multivariate benchmarks. Each target trajectory is the output of an AC power-flow solve, and PowerPhase ships with constraint-aware metrics (Safety\_mBrier, NECV, $\mathrm{CVaR}_\alpha$) that complement CRPS and Distortion. Across eight baselines and three seeds, distributional accuracy and constraint satisfaction rank models differently, a trade-off we term \emph{safety--fidelity}. We further propose PowerForge, a scenario-based quantile forecaster with type-specific decoding heads and a causal bridge between variable groups, which achieves the best average rank on every grid.

\end{abstract}

\section{Introduction}
\label{sec:intro}

Multivariate time series forecasting underlies decision-making in
domains as diverse as retail demand, transportation, climate,
healthcare, and finance \citep{benidis2022deep, liu2026falconxtimeseriesfoundation, ansari2025chronos2univariateuniversalforecasting, dai2026positionuniversaltimeseries}. Forecasting on power systems has become one of the most operationally consequential instances of this problem \citep{hong2020energy}, because short-term predictions of grid state feed into reserve scheduling, contingency screening, and short-term market clearing \citep{morales2013integrating, xu2025optimal,OPFormer}. As renewables and distributed loads push grids toward more stochastic operation \citep{hong2016probabilistic,zhang2025disturbed}, probabilistic forecasts of grid state have moved from a research curiosity to a routine operational tool \citep{roald2023power}.

Yet forecasting on a transmission grid is more constrained than the
classic multivariate setting. At every bus, an operator observes
four physically coupled quantities: the active power $P$, the
reactive power $Q$, the voltage magnitude $V$, and the voltage angle
$\theta$ \citep{glover2012power}. These quantities are tied together through the AC power
flow equations \citep{wood2013power}, so a useful forecast must
respect not only temporal dependencies across channels but also
instantaneous physical feasibility. \textit{This creates two difficulties
that existing benchmarks and models do not address}. \textbf{First,} standard
probabilistic scoring rules such as CRPS and log-likelihood average
over the full forecast distribution and do not distinguish whether
errors fall inside or outside the operating envelope. A model that
achieves low CRPS while systematically underestimating voltage
excursions can be more dangerous in practice than one with slightly
worse distributional fit but reliable constraint satisfaction.
\textbf{Second,} the number of channels grows linearly with the network,
which means a moderately sized transmission system already exceeds
the scale at which most public multivariate benchmarks live.

Existing public benchmarks reflect this gap. The multivariate suites commonly used in the machine learning literature, including ETT \citep{zhou2021informer}, Electricity, Traffic \citep{lai2018modeling}, and the Wikipedia Web Traffic series in the multivariate configuration of \citet{salinas2019high}, top out at roughly $2{,}000$ jointly forecasted channels. None of these datasets distinguishes channels by physical role or ships with a feasibility model, so there is no standard way to measure whether a probabilistic forecast respects the physical constraints of the system it describes. The power-systems community has developed the opposite kind of artifact \citep{chassin2014gridlab}. Tools such as pandapower \citep{thurner2018pandapower} and the PEGASE test cases \citep{josz2016ac} expose network models with thousands of buses \citep{meinecke2020simbench}, but they are simulators rather than benchmarks. They provide neither standardized forecasting splits nor probabilistic metrics, and they are typically used without strong learning baselines. Existing bridges between the two communities cover either snapshot regression or small-scale temporal data, leaving the transmission-scale probabilistic regime unaddressed.

On the modeling side, multivariate probabilistic forecasting has
progressed quickly, but largely on smaller settings. Autoregressive
likelihood models such as DeepAR \citep{salinas2020deepar},
conditioned flow models such as Transformer-TempFlow
\citep{rasul2020multivariate}, diffusion approaches such as TimeGrad
\citep{rasul2021autoregressive}, and copula-based models such as
TACTiS \citep{drouin2022tactis} all rely on per-step density
estimation, which scales poorly when the number of channels reaches
tens of thousands. A complementary line of work replaces density
estimation with a small set of weighted scenarios, as in TimeMCL
\citep{cortes2025winnertakesall} and TimePrism \citep{dai_samples_2025}. This
scenario-based view is well aligned with how grid operators already
reason about uncertainty in terms of finite contingencies, but it
has not yet been pushed to transmission scale.

We address both the benchmarking gap and the modeling gap. On the
modeling side, we propose \textbf{PowerForge}, a probabilistic forecasting model whose design is driven by the constraints that
transmission-scale grids impose. To evaluate PowerForge and future
methods under these constraints, we release \textbf{PowerPhase},
a companion benchmark. Our contributions are:
\paragraph{Contributions.}
\begin{itemize}[leftmargin=1.5em,itemsep=2pt,topsep=2pt]
    \item We release \textbf{PowerPhase}, a multivariate probabilistic forecasting benchmark covering six transmission networks of up to $36{,}964$ physically typed channels, more than an order of magnitude beyond the largest canonical multivariate benchmark (Wiki, $2{,}000$ channels), bringing the high-dimensional regime into reach of standard forecasting evaluation.

    \item PowerPhase introduces voltage-safety evaluation into the protocol. Each series is generated by AC power
    flow on perturbed load and generation profiles, and the benchmark
    ships with a constraint-aware metric suite
    (Safety\_mBrier,  NECV, $\mathrm{CVaR}_\alpha$) that
    complements average-case scores such as CRPS, so models are
    assessed on voltage safety in addition to distributional
    accuracy. Evaluation across five grid sizes, three seeds, and eight
baselines reveals that distributional accuracy and constraint
satisfaction rank models differently, a trade-off we term
\emph{safety--fidelity}.

\item We propose \textbf{PowerForge}, a scenario-based probabilistic model designed for this regime. It predicts a small set of ordered quantile scenarios that keep training and inference tractable at tens of thousands of channels. Three architectural priors drive the design: an anchor-based residual representation that removes the dominant diurnal component before encoding, type-aware decoding heads for channels with structurally different supports, and a causal cross-type bridge that encodes a known directional dependence as an architectural prior. Training uses an ordered-quantile objective with physics terms that remain tractable at the full PowerPhase scale.

\end{itemize}

\section{Related Work}
\label{sec:related_work}

\paragraph{Multivariate probabilistic forecasting.}
Approaches to multivariate probabilistic forecasting differ in how they parameterize the predictive distribution \citep{gneiting2014probabilistic}. DeepAR \citep{salinas2020deepar} learns an autoregressive parametric likelihood, deep state-space models combine recurrent latent dynamics with
probabilistic emissions~\citep{rangapuram2018deep}, and \citet{salinas2019high} extends this idea with a low-rank Gaussian copula. Normalizing-flow variants such as TempFlow and Transformer-TempFlow \citep{rasul2020multivariate} replace the parametric head with a conditioned flow, and TimeGrad \citep{rasul2021autoregressive} and ScoreGrad \citep{yan2021scoregrad} parameterize the per-step conditional through denoising diffusion or score matching. Attentional copula models go further by factoring the joint distribution through an explicit copula transformer, as in TACTiS \citep{drouin2022tactis} and TACTiS-2 \citep{ashok2023tactis}. Instead, scenario-based line of work, including TimeMCL \citep{cortes2025winnertakesall} and TimePrism \citep{dai_samples_2025}, predicts a small set of weighted hypotheses.  How these approaches behave when channel numbers are in the tens of thousands remains untested.

\paragraph{Multivariate time series benchmarks.}
The methods above are typically validated on a series of public benchmarks. Commonly used datasets in time series, like ETT, Electricity, Traffic, Solar, Exchange-Rate, and Wiki datasets by \citet{lai2018modeling, salinas2019high, zhou2021informer, wuAutoformerDecompositionTransformers2021b} push channel counts up to roughly $2{,}000$.  Related high-dimensional forecasting work has
explored global-local architectures~\citep{sen2019think}, but not in
transmission-scale grid-state benchmarks. Aggregated repositories such as the Monash Forecasting Archive \citep{godahewa2021monash}, the GluonTS benchmark suite \citep{alexandrov2020gluonts}, and LOTSA \citep{woo2024unified} broaden coverage across domains but do not focus on per-dataset dimensionality. Meanwhile, these benchmarks have not exposed the physical structure required for probabilistic forecasters at the scale of an actual transmission grid \citep{xu2026fideltshighfidelitymultimodalbenchmark}.

\paragraph{Power-system data for machine learning.}
Machine learning benchmarks derived from power grids have grown rapidly \citep{varbella2024powergraph}, but each covers only part of the evaluation. \textbf{One group} supplies physical models without temporal structure \citep{babaeinejadsarookolaee2021pglib}: PF$\Delta$ \citep{rivera2025pfdelta} benchmarks steady-state power-flow regression up to $2{,}000$ buses, and OPFData \citep{lovett2024opfdata} provides collections of optimal-power-flow instances. Both offer data snapshots in an independent operating point with no temporal ordering, so neither can be used for time-series evaluation. \textbf{A second group} provides temporal signals but lacks the scale or the physical evaluation that transmission grids demand. PSML \citep{zheng2022psml} builds multi-scale time series from a joint transmission-distribution co-simulation on a 23-bus transmission network and 13-bus distribution feeders, focusing on cross-scale dynamics rather than the transmission-scale, bus-level probabilistic forecasting with AC-feasibility-aware evaluation considered here. Real PMU recordings aggregated through Open Power System Data \citep{wiese2019open} are partially observable and carry no feasibility ground truth. \textbf{PowerPhase is, to our knowledge, the first benchmark that combines transmission-scale channel counts, physically typed per-bus variables, temporal trajectory structure, and probabilistic evaluation with constraint-aware metrics.}

\paragraph{Risk-aware and constraint-aware evaluation.}
On the training side, risk-aware losses based on conditional value-at-risk \citep{rockafellar2000cvar} and distributionally robust objectives \citep{rahimian2019dro} reweight tail behaviour so that the model pays more attention to extreme outcomes, and physics-informed approaches \citep{raissi2019pinn, dai_socgate_2025, dai_socnet_2025} encode governing equations as soft regularizers to encourage physically consistent predictions. On the evaluation side, conformal prediction \citep{angelopoulos2021gentle} and its time-series extensions \citep{xu2023sequential} offer marginal coverage guarantees but do not measure the magnitude of constraint violations when coverage fails. Across both groups, results are reported on datasets with standard distributional metrics with limited benchmark-level assessment on how methods satisfy physical constraints.

\section{The PowerPhase Benchmark}
\label{sec:powerphase}



\subsection{Generation Procedure}
\label{sec:pipeline}

\paragraph{Source signals and networks.}
We use German TSO data from Open Power System Data
\citep{wiese2019open} covering 2015 and 2016 at $15$-minute
resolution, retaining the aggregate load, solar generation,
and wind generation series after linear interpolation of
missing entries. The resulting record has $70{,}176$ time
steps. We instantiate six standard pandapower test cases
\citep{thurner2018pandapower} ranging from $500$ to $9{,}241$
buses, listed with per-network citations in
Table~\ref{tab:instances}.

\begin{figure}[t]
    \centering
    \includegraphics[width=\linewidth]{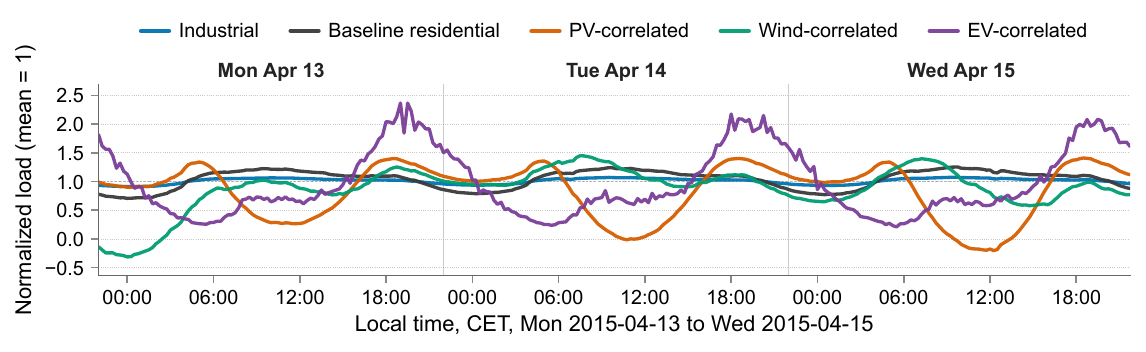}
    \caption{The five daily-shape load profiles used in the
    per-node injection synthesis, shown over three consecutive
    days in April 2015 and normalised to unit mean. Industrial
    profiles are nearly flat, baseline residential follows a
    morning-evening double peak, PV- and wind-correlated
    profiles track renewable availability, and EV-correlated
    profiles peak in the late evening. Each load bus is
    assigned one profile at instantiation; high-voltage buses
    are preferentially assigned industrial profiles.}
    \label{fig:profiles}
\end{figure}

\paragraph{Per-node injection synthesis.}
Each load bus is assigned one of five daily-shape profiles
(Figure~\ref{fig:profiles}) and a per-node power factor;
spatially correlated regional noise is added before the
power-flow solve. Profile assignment rules, noise levels,
and power-factor ranges are detailed in
Appendix~\ref{app:pipeline}.

\paragraph{AC power-flow solve.}
For every time step we run AC power flow through pandapower
using Newton--Raphson with iterative back-off on the injection
magnitudes until convergence. Failed steps, which account for
fewer than $1\%$ of the record on every network, are
forward-filled with the last converged state. The final output
for each bus is the four-vector $(P, Q, V, \theta)$.

\subsection{Forecasting Task }
\label{sec:task}
A network with $N$ buses produces a multivariate signal
$Z_t \in \mathbb{R}^{4N}$ ordered as
$[P_n, Q_n, V_n, \theta_n]_{n=1}^{N}$. Given a context of
length $T_h = 672$, the task is to model the joint predictive
distribution over $Z_{t+1:t+T_p}$ with $T_p = 96$,
corresponding to a one-day horizon at $15$-minute resolution.
Evaluation uses rolling-origin testing with ten equally spaced
prediction windows per network. Normalisation conventions for
each variable type are given in
Appendix~\ref{app:normalisation}.

\subsection{Metric Suite}
\label{sec:metrics}
\begin{table}[t]
    \centering
    \small
    \caption{The six networks in PowerPhase. \emph{PQ} is the number of
    load buses; the remaining buses are voltage-controlled (PV) or the
    reference bus. \emph{Channels} equals $4 \times \text{buses}$.
    \emph{Steps} is the total length at $15$-minute resolution;
    \emph{Test} is the number of rolling-origin prediction windows
    per network.}
    \label{tab:instances}
    \begin{tabular}{lrrrrrr}
    \toprule
    Network & Buses & PQ & Channels & Steps & Pred. & Test \\
    \midrule
    ACTIVSg\,500 \citep{birchfield2017grid}      & $500$     & $410$     & $2{,}000$   & $70{,}176$ & $96$ & $10$ \\
    PEGASE\,1354 \citep{josz2016ac}              & $1{,}354$ & $1{,}094$ & $5{,}416$   & $70{,}176$ & $96$ & $10$ \\
    Polish\,2383 \citep{zimmerman2010matpower}   & $2{,}383$ & $2{,}056$ & $9{,}532$   & $70{,}176$ & $96$ & $10$ \\
    PEGASE\,2869 \citep{josz2016ac}              & $2{,}869$ & $2{,}359$ & $11{,}476$  & $70{,}176$ & $96$ & $10$ \\
    Polish\,3120 \citep{zimmerman2010matpower}   & $3{,}120$ & $2{,}771$ & $12{,}480$ & $70{,}176$ & $96$ & $10$ \\
    PEGASE\,9241 \citep{josz2016ac}              & $9{,}241$ & $7{,}796$ & $\mathbf{36{,}964}$ & $70{,}176$ & $96$ & $10$ \\
    \bottomrule
    \end{tabular}
\end{table}

We score forecasters along two axes. Statistical fidelity
uses CRPS \citep{gneiting2007strictly} averaged over channels
and time, plus Distortion \citep{cortes2025winnertakesall}
for scenario models. Definitions and the
probability-weighted CRPS variant \citep{hersbach2000decomposition} are deferred to
Appendix~\ref{app:metrics}.

Operational voltage safety is evaluated on the voltage band
$[V_{\min},V_{\max}]=[0.95,1.05]\,\mathrm{p.u.}$.
For each voltage-channel evaluation point $(t,d)\in\mathcal{V}$,
let
$Y_{t,d}=\mathbf{1}\{V_{t,d}\notin[V_{\min},V_{\max}]\}$ and
$\hat Y_{t,d}^{(k)}=\mathbf{1}\{\hat V_{t,d}^{(k)}
\notin[V_{\min},V_{\max}]\}$.
We also define the scenario violation magnitude
\[
\delta_{t,d}^{(k)}
=(V_{\min}-\hat V_{t,d}^{(k)})_{+}
+(\hat V_{t,d}^{(k)}-V_{\max})_{+}.
\]
For scenario weights $w_k$ (uniform $1/K$ for sample-based methods),
we report
\begin{align}
\mathrm{Safety\_mBrier}
&=\frac{1}{|\mathcal V|}
\sum_{(t,d)\in\mathcal V}\sum_k
w_k\bigl(\hat Y_{t,d}^{(k)}-Y_{t,d}\bigr)^2,
\label{eq:safety_mbrier}\\
\mathrm{NECV}
&=\frac{1}{|\mathcal V|}
\sum_{(t,d)\in\mathcal V}\sum_k
w_k\,\delta_{t,d}^{(k)},
\label{eq:necv}\\
\mathrm{CVaR}_{0.1}
&=\frac{1}{|\mathcal V|}
\sum_{(t,d)\in\mathcal V}
\frac{1}{\lceil0.1K\rceil}
\sum_{k\in\mathcal T_{0.1}(t,d)}
\delta_{t,d}^{(k)} ,
\label{eq:cvar}
\end{align}
where $\mathcal{T}_{0.1}(t,d)$ contains the $\lceil0.1K\rceil$
scenarios with the largest $\delta_{t,d}^{(k)}$. Safety\_mBrier
measures violation detection, whereas NECV and $\mathrm{CVaR}_{0.1}$
measure average and worst-decile violation severity. Additional details, including the relation to the standard Brier score \citep{murphy1973new}, are given in
Appendix~\ref{app:metrics}.


\section{The PowerForge Model}
\label{sec:powerforge}
\subsection{Overview}
\label{sec:powerforge_overview}

\begin{figure}[t]
    \centering
    \includegraphics[width=\linewidth]{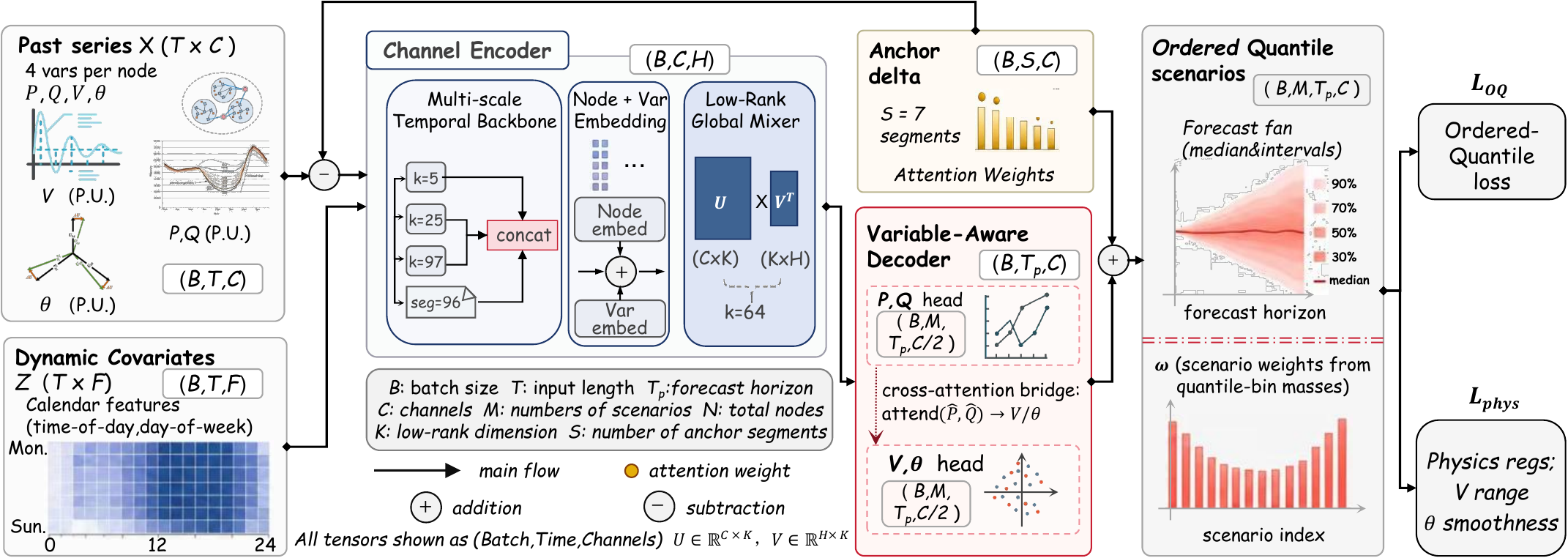}
    \caption{PowerForge architecture. The past series $X \in \mathbb{R}^{T \times C}$ is anchor-subtracted and passed through a channel encoder (multi-scale temporal backbone with kernels $\{5, 25, 97\}$, node and variable embeddings, and a low-rank global mixer). Calendar covariates $Z$ are injected as time-aggregated context. A variable-aware decoder produces $M$ scenario branches with separate $P, Q$ and $V, \theta$ heads linked by a causal cross-attention bridge from $\hat P, \hat Q$ to $V, \theta$. The anchor is added back, and pointwise sorting along the scenario axis yields ordered quantile scenarios with quantile-bin weights $\omega$. The model is trained with an ordered-quantile pinball loss $\mathcal{L}_{\mathrm{OQ}}$ and physics regularisers $\mathcal{L}_{\mathrm{phys}}$.}
    \label{fig:architecture}
\end{figure}

The PowerForge architecture is shaped by three constraints of the PowerPhase regime. First, cross-variable interaction must be sub-quadratic in the channel count, since $C$ on the larger PowerPhase grids reaches $10^4$. Second, decoding must be type-heterogeneous, since the four physical quantities have structurally different supports. Third, when the variable types carry a known directional dependence, the architecture should reflect that structure rather than discover it from gradients.

Figure~\ref{fig:architecture} traces these principles through the model. The past series $X \in \mathbb{R}^{T \times C}$ is first projected into a residual space by subtracting an anchor $Z^{\mathrm{ref}}$ that captures the dominant diurnal pattern (\S\ref{sec:powerforge_anchor}). The residual is then consumed by a channel encoder and a low-rank global mixer that exchanges information across all $C$ channels through $K \ll C$ shared tokens (\S\ref{sec:powerforge_mixer}). A variable-aware decoder (\S\ref{sec:powerforge_decoder}) emits $M$ scenario branches through separate $P, Q$ and $V, \theta$ heads, with a causal cross-attention bridge that conditions $V, \theta$ on the predicted $\hat P, \hat Q$ and reflects the AC power-flow direction. Adding the anchor back and sorting along the scenario axis produces ordered quantile trajectories $\{(\hat Z^{(m)}, w_m)\}_{m=1}^{M}$. With $M$ in the low tens, all scenarios are produced in a single forward pass, in contrast to the per-step sampling of autoregressive density baselines. Training uses an ordered-quantile pinball loss with physics regularisers (\S\ref{sec:powerforge_training}).

\subsection{Reference-Anchored Residual Space}
\label{sec:powerforge_anchor}
The model operates in a residual space defined relative to a learned reference $Z^{\mathrm{ref}} \in \mathbb{R}^{L \times C}$. The reference is subtracted from the input history before encoding, and added back to the decoder output to recover absolute predictions. Working in this space removes the dominant diurnal component shared across all power-grid signals, leaving the encoder and decoder to model only the deviations.

The reference is built from the recent history. The last $S$ daily segments of length $L$ are extracted, and each channel forms its own reference as a per-channel attention-weighted average. Let $X^{(s)}_c$ denote the $s$-th segment of channel $c$ and $X^{\mathrm{query}}_c$ the last $Q$ steps of the input. Then
\begin{equation}
    Z^{\mathrm{ref}}_c
    = \sum_{s=1}^{S} a_{s,c} \, X^{(s)}_c,
    \qquad
    a_{s,c}
    = \frac{\exp\!\bigl(\rho(X^{(s)}_c, X^{\mathrm{query}}_c) / \tau\bigr)}
           {\sum_{s'=1}^{S} \exp\!\bigl(\rho(X^{(s')}_c, X^{\mathrm{query}}_c) / \tau\bigr)},
    \label{eq:anchor}
\end{equation}
where $\rho$ is the Pearson correlation between the last $Q$ steps of $X^{(s)}_c$ and $X^{\mathrm{query}}_c$, computed per channel so different channels can favour different segments. A short bias correction further aligns the reference to the most recent observations near the input boundary. Details are in Appendix~\ref{app:anchor_delta}.

\subsection{Channel Encoder and Global Token Mixer}
\label{sec:powerforge_mixer}

The channel encoder (Appendix~\ref{app:encoder}) maps the residual input into a per-channel representation $h \in \mathbb{R}^{B \times C \times H}$ via a multi-scale temporal backbone with kernel sizes $\{5, 25, 97\}$ together with node, variable, and calendar embeddings. Self-attention across channels would scale as $\mathcal{O}(C^2)$, which is prohibitive at this scale \citep{wang2020linformer}. We replace this dense interaction with a low-rank mixer that maintains $K$ learnable global tokens $G \in \mathbb{R}^{K \times H}$ and alternates read and write attention. Channel states first attend to $G$ (read), $G$ then attends back to the updated channels (write), and a second read propagates the refreshed tokens to the channels, with the two read steps sharing projection weights. Both directions cost $\mathcal{O}(CK)$ with $K \ll C$, so the total cost is linear in $C$. The $K$ tokens act as a rank-$K$ bottleneck on cross-channel interaction, which is well matched to grids where system-wide load and renewable generation mix a handful of latent factors into thousands of channels.

\subsection{Type-Aware Decoder with Cross-Type Bridge}
\label{sec:powerforge_decoder}

The decoder operates in the residual space defined in \S\ref{sec:powerforge_anchor}, producing $M$ hypothesis trajectories through per-type output heads and a causal cross-type bridge, instantiated on PowerPhase with $\{P, Q, V, \theta\}$.

\paragraph{Type-specific heads.}
Each channel state $h_c$ is fused with $M$ learnable scenario embeddings through the additive injection $h^{(m)}_c = h_c + e_m$, so the $M$ branches differ at the embedding level before any per-type processing. A per-type head then maps each fused state to a residual prediction $\delta^{(m)}_c$, with the parameterisation chosen to match the support of each variable. The $P$ and $Q$ heads emit an unbounded linear projection of the fused state, with a small Gaussian perturbation added as a stochastic regulariser. The $V$ and $\theta$ heads pass a learned projection through a $\tanh$ gate scaled by a per-type magnitude $\Delta_c$, matching the narrow dynamic range these channels occupy after anchor subtraction. Full head equations and angular wrapping for $\theta$ are given in Appendix~\ref{app:heads}.

\paragraph{Causal cross-type bridge.}
The AC power-flow equations determine bus voltages and angles given the active and reactive power injections \citep{wood2013power}. The decoder reflects this direction by producing $P$ and $Q$ predictions first, projecting each predicted trajectory into a hidden token $\phi^{(u)}_c$ for $u \in \{P, Q\}$, and letting the $V$ and $\theta$ hidden states attend to them through scaled dot-product attention,
\begin{equation}
    h^{(m, t)}_c \leftarrow h^{(m, t)}_c + \gamma \cdot \mathrm{Attn}\!\bigl(W_q\, h^{(m, t)}_c,\; W_k\, \Phi^{(m)}_c,\; W_v\, \Phi^{(m)}_c\bigr), \quad t \in \{V, \theta\},
    \label{eq:bridge}
\end{equation}
where $\Phi^{(m)}_c = \{\phi^{(P)}_c, \phi^{(Q)}_c\}$ collects the two source tokens, $W_q, W_k, W_v$ are projections shared across nodes and scenarios, and $\gamma$ is a scale initialised at $1$. The bridge is an inductive bias rather than a hard constraint, since the projection weights can downweight the attention output when the conditioning is uninformative.

\subsection{Training Objective}
\label{sec:powerforge_training}

After the per-type heads emit residuals and the anchor is added back, the $M$ trajectories are pointwise sorted along the scenario axis, so branch $m$ is interpreted as the quantile estimator at level $\tau_m$, with $\tau_1 < \cdots < \tau_M$ controlled by a learnable $\mathrm{Beta}(\alpha, \beta)$ shape. Each branch is supervised by a pinball loss at its assigned level \citep{koenker1978regression},
\begin{equation}
    \mathcal{L}_{\mathrm{OQ}} = \frac{1}{B M T_p C} \sum_{b, m, t, c} \rho_{\tau_m}\!\bigl(Z_{b,t,c} - \hat Z^{(m)}_{b,t,c}\bigr), \qquad \rho_\tau(u) = u\bigl(\tau - \mathbf{1}\{u < 0\}\bigr).
    \label{eq:pinball}
\end{equation}
where $B$ is the batch size and the sum runs over the $T_p$ forecast steps. Four learnable per-type coefficients (one each for $P, Q, V, \theta$) scale the per-type contributions to balance gradients across variables, initialised from inverse volatility on a small set of calibration batches. A physics regulariser dampens the voltage-residual magnitude and penalises temporal discontinuities in the angle channel. Scenario weights $w_m$ derived from the $\tau_m$ grid (Appendix~\ref{app:training}) are used at evaluation in weighted CRPS, weighted Safety\_mBrier, and the categorical sampler. Quantile-fan regularisers and the Beta prior are detailed in Appendix~\ref{app:training}.

\section{Experiments}
\label{sec:experiments}
\subsection{Setup}
\label{sec:setup}

\paragraph{Baselines and Protocol.}
We compare PowerForge against eight probabilistic forecasting
baselines spanning three families.
\emph{Sample-based density models}: DeepAR
\citep{salinas2020deepar}, TempFlow and Transformer-TempFlow
\citep{rasul2020multivariate}, TimeGrad
\citep{rasul2021autoregressive}, and TACTiS-2
\citep{ashok2023tactis}.
\emph{Scenario-based models}: TimeMCL
\citep{cortes2025winnertakesall} and TimePrism \citep{dai_samples_2025}.
\emph{Statistical baseline}: ETS (exponential smoothing),
fitted independently per channel. Every model is trained on each of the five main PowerPhase networks (ACTIVSg\,500 through Polish\,3120) with three random seeds, and we report the mean across seeds. At the upper end of the benchmark, PEGASE\,9241 ($36{,}964$ channels) is evaluated against four baselines spanning scenario-based, conditioned-flow, and statistical families (Appendix~\ref{app:9241}). Each forecasting window uses $T_h = 672$ context steps and predicts $T_p = 96$ steps ahead, evaluated through rolling-origin testing with ten windows per network. Density baselines draw $K = 100$ forecast samples at test time, and scenario-based models produce weighted hypothesis trajectories per their respective configurations (PowerForge: $M = 16$). Train, validation, and test splits are fixed across all models and seeds (Appendix~\ref{app:splits}).
All deep models are trained on a single NVIDIA RTX PRO 6000 Blackwell GPU with Adam, and the checkpoint with the lowest
validation loss is used for evaluation. Deep baselines follow 
their GluonTS \citep{alexandrov2020gluonts} reference configurations. 
ETS is fitted per channel by maximum likelihood on CPU. Full PowerForge 
and baseline hyperparameters are in Appendix~\ref{app:implementation}.

\begin{table*}[t]
\caption{Probabilistic forecasting results on \textsc{PowerPhase} (three representative grids; full results in Appendix~\ref{app:full_table}).
CRPS and Distortion measure \emph{fidelity}; Safety\_mBrier and $\text{CVaR}_{0.1}$ measure \emph{safety}.
All values are mean{\scriptsize$\pm$std} over three seeds.
\textbf{Bold}: best; \underline{underline}: second best.
\emph{Rank} is the model's average rank within the grid across the four metrics.}
\label{tab:main}
\centering
\setlength{\tabcolsep}{3.5pt} 
\small
\begin{tabular}{@{}l l  cc  cc  c@{}}
\toprule
 & & \multicolumn{2}{c}{\emph{Fidelity}\,$\downarrow$}
   & \multicolumn{2}{c}{\emph{Safety}\,$\downarrow$}
   & \\
\cmidrule(lr){3-4} \cmidrule(lr){5-6}
\textbf{Grid} & \textbf{Model}
  & CRPS & Distortion
  & Safety\_mBrier & $\text{CVaR}_{0.1}$
  & Rank \\
\midrule
  500-bus
  & DeepAR
    & 0.1156{\scriptsize$\pm$.0240}
    & 0.5537{\scriptsize$\pm$.1310}
    & 0.9519{\scriptsize$\pm$.0233}
    & 2.1753{\scriptsize$\pm\phantom{0}$.6901}
    & 8.8 \\
  
  & ETS
    & 0.0064{\scriptsize$\pm$.0000}
    & 0.0481{\scriptsize$\pm$.0000}
    & 0.0561{\scriptsize$\pm$.0002}
    & \underline{0.0116}{\scriptsize$\pm\phantom{0}$.0000}
    & 4.2 \\
  
  & TimeMCL
    & \underline{0.0054}{\scriptsize$\pm$.0003}
    & \underline{0.0118}{\scriptsize$\pm$.0001}
    & \underline{0.0200}{\scriptsize$\pm$.0200}
    & 0.1906{\scriptsize$\pm\phantom{0}$.1907}
    & 3.2 \\
  
  & TimePrism
    & 0.0077{\scriptsize$\pm$.0003}
    & 0.0195{\scriptsize$\pm$.0023}
    & 0.0221{\scriptsize$\pm$.0062}
    & 2.9950{\scriptsize$\pm$1.1445}
    & 5.5 \\
  
  & TACTiS-2
    & 0.0057{\scriptsize$\pm$.0002}
    & 0.0139{\scriptsize$\pm$.0006}
    & \textbf{0.0000}{\scriptsize$\pm$.0000}
    & \textbf{0.0000}{\scriptsize$\pm\phantom{0}$.0000}
    & \underline{2.2} \\
  
  & TempFlow
    & 0.0144{\scriptsize$\pm$.0008}
    & 0.0679{\scriptsize$\pm$.0005}
    & 0.2740{\scriptsize$\pm$.0519}
    & 0.0332{\scriptsize$\pm\phantom{0}$.0043}
    & 5.5 \\
  
  & Trans-TempFlow
    & 0.0146{\scriptsize$\pm$.0006}
    & 0.0702{\scriptsize$\pm$.0011}
    & 0.2896{\scriptsize$\pm$.0581}
    & 0.0380{\scriptsize$\pm\phantom{0}$.0100}
    & 6.5 \\
  
  & TimeGrad
    & 0.0160{\scriptsize$\pm$.0011}
    & 0.0954{\scriptsize$\pm$.0092}
    & 0.7630{\scriptsize$\pm$.0192}
    & 0.3316{\scriptsize$\pm\phantom{0}$.0358}
    & 7.8 \\
  
  & \textbf{PowerForge (Ours)}
    & \textbf{0.0030}{\scriptsize$\pm$.0008}
    & \textbf{0.0072}{\scriptsize$\pm$.0019}
    & \textbf{0.0000}{\scriptsize$\pm$.0000}
    & \textbf{0.0000}{\scriptsize$\pm\phantom{0}$.0000}
    & \textbf{1.2} \\
\midrule
  2383-bus
  & DeepAR
    & 0.0583{\scriptsize$\pm$.0261}
    & 0.4124{\scriptsize$\pm$.1674}
    & 0.5440{\scriptsize$\pm$.1632}
    & 0.1831{\scriptsize$\pm\phantom{0}$.1303}
    & 8.2 \\
  
  & ETS
    & 0.0111{\scriptsize$\pm$.0000}
    & 0.1513{\scriptsize$\pm$.0000}
    & 0.0352{\scriptsize$\pm$.0001}
    & 0.0099{\scriptsize$\pm\phantom{0}$.0000}
    & 6.5 \\
  
  & TimeMCL
    & 0.0070{\scriptsize$\pm$.0008}
    & \underline{0.0147}{\scriptsize$\pm$.0016}
    & 0.0109{\scriptsize$\pm$.0068}
    & 0.0322{\scriptsize$\pm\phantom{0}$.0547}
    & 4.5 \\
  
  & TimePrism
    & 0.0096{\scriptsize$\pm$.0085}
    & 0.0250{\scriptsize$\pm$.0164}
    & 0.0294{\scriptsize$\pm$.0255}
    & 3.5007{\scriptsize$\pm$1.1821}
    & 6.8 \\
  
  & TACTiS-2
    & \textbf{0.0039}{\scriptsize$\pm$.0000}
    & 0.0188{\scriptsize$\pm$.0000}
    & 0.0036{\scriptsize$\pm$.0000}
    & 0.0007{\scriptsize$\pm\phantom{0}$.0000}
    & 3.5 \\
  
  & TempFlow
    & 0.0057{\scriptsize$\pm$.0001}
    & 0.0153{\scriptsize$\pm$.0026}
    & \underline{0.0032}{\scriptsize$\pm$.0001}
    & \underline{0.0005}{\scriptsize$\pm\phantom{0}$.0000}
    & \underline{2.8} \\
  
  & Trans-TempFlow
    & 0.0060{\scriptsize$\pm$.0002}
    & 0.0174{\scriptsize$\pm$.0002}
    & \underline{0.0032}{\scriptsize$\pm$.0000}
    & \underline{0.0005}{\scriptsize$\pm\phantom{0}$.0000}
    & 3.2 \\
  
  & TimeGrad
    & 0.0169{\scriptsize$\pm$.0016}
    & 0.1753{\scriptsize$\pm$.0826}
    & 0.5454{\scriptsize$\pm$.0505}
    & 0.8028{\scriptsize$\pm\phantom{0}$.4729}
    & 8.2 \\
  
  & \textbf{PowerForge (Ours)}
    & \underline{0.0042}{\scriptsize$\pm$.0019}
    & \textbf{0.0084}{\scriptsize$\pm$.0025}
    & \textbf{0.0010}{\scriptsize$\pm$.0002}
    & \textbf{0.0004}{\scriptsize$\pm\phantom{0}$.0000}
    & \textbf{1.2} \\
\midrule
  3120-bus
  & DeepAR
    & 0.1408{\scriptsize$\pm$.0148}
    & 0.7666{\scriptsize$\pm$.0842}
    & 0.7883{\scriptsize$\pm$.0160}
    & 1.6903{\scriptsize$\pm\phantom{0}$.5156}
    & 8.8 \\
  
  & ETS
    & 0.0180{\scriptsize$\pm$.0001}
    & 0.1954{\scriptsize$\pm$.0000}
    & 0.0987{\scriptsize$\pm$.0001}
    & 0.0225{\scriptsize$\pm\phantom{0}$.0000}
    & 6.8 \\
  
  & TimeMCL
    & 0.0092{\scriptsize$\pm$.0052}
    & \underline{0.0162}{\scriptsize$\pm$.0044}
    & 0.1100{\scriptsize$\pm$.1071}
    & 0.2240{\scriptsize$\pm\phantom{0}$.2385}
    & 5.2 \\
  
  & TimePrism
    & 0.0083{\scriptsize$\pm$.0018}
    & 0.0213{\scriptsize$\pm$.0119}
    & 0.0579{\scriptsize$\pm$.0383}
    & 5.4230{\scriptsize$\pm$3.2742}
    & 6.0 \\
  
  & TACTiS-2
    & \underline{0.0043}{\scriptsize$\pm$.0000}
    & 0.0297{\scriptsize$\pm$.0001}
    & 0.0428{\scriptsize$\pm$.0006}
    & 0.0071{\scriptsize$\pm\phantom{0}$.0000}
    & 4.0 \\
  
  & TempFlow
    & 0.0055{\scriptsize$\pm$.0001}
    & 0.0178{\scriptsize$\pm$.0001}
    & 0.0187{\scriptsize$\pm$.0000}
    & 0.0035{\scriptsize$\pm\phantom{0}$.0000}
    & 3.5 \\
  
  & Trans-TempFlow
    & 0.0054{\scriptsize$\pm$.0002}
    & 0.0175{\scriptsize$\pm$.0006}
    & \underline{0.0184}{\scriptsize$\pm$.0004}
    & \underline{0.0034}{\scriptsize$\pm\phantom{0}$.0000}
    & \underline{2.5} \\
  
  & TimeGrad
    & 0.0123{\scriptsize$\pm$.0069}
    & 0.1295{\scriptsize$\pm$.0503}
    & 0.4234{\scriptsize$\pm$.1359}
    & 0.5695{\scriptsize$\pm\phantom{0}$.2468}
    & 7.2 \\
  
  & \textbf{PowerForge (Ours)}
    & \textbf{0.0038}{\scriptsize$\pm$.0011}
    & \textbf{0.0076}{\scriptsize$\pm$.0010}
    & \textbf{0.0068}{\scriptsize$\pm$.0000}
    & \textbf{0.0025}{\scriptsize$\pm\phantom{0}$.0000}
    & \textbf{1.0} \\
\bottomrule
\end{tabular}
\end{table*}


\subsection{Main Result}
\label{exp1}

Table~\ref{tab:main} reports the four-metric scoreboard on three representative grids. The results on all five main grids
are in Appendix~\ref{app:full_table}, and the result of
PEGASE\,9241 evaluation is in Appendix~\ref{app:9241}. PowerForge attains the best average rank on every grid ($1.2$, $1.2$, $1.0$) and is the only model that stays in the top tier across both fidelity and Safety axes. At 500-bus, it achieves the best score on all four metrics, with Safety\_mBrier and $\mathrm{CVaR}_{0.1}$ tied at $0$ alongside TACTiS-2. At 2383-bus, PowerForge matches TACTiS-2 on CRPS within one standard deviation, while attaining the best Distortion and the lowest Safety\_mBrier ($0.0010$, vs.\ $0.0032$ for TempFlow and Transformer-TempFlow) and $\mathrm{CVaR}_{0.1}$. At 3120-bus, it is strict best on every metric, with a Distortion more than $2\times$ lower than the next baseline.

The table makes the safety--fidelity trade-off concrete. TimePrism is the clearest instance, with a competitive CRPS at 2383-bus ($0.0096$) but the largest $\mathrm{CVaR}_{0.1}$ ($3.50$) of any deep baseline, indicating that the winner-takes-all training criterion produces low-probability hypotheses that drift far outside the operational band. TimeMCL shows a milder version of the same pattern at 500-bus, where strong CRPS ($0.0054$) coexists with mediocre $\mathrm{CVaR}_{0.1}$ ($0.19$). DeepAR fails on both axes, with the worst CRPS on every grid and Safety\_mBrier never below $0.54$.
\begin{figure}[t]
    \centering
    \includegraphics[width=\textwidth]{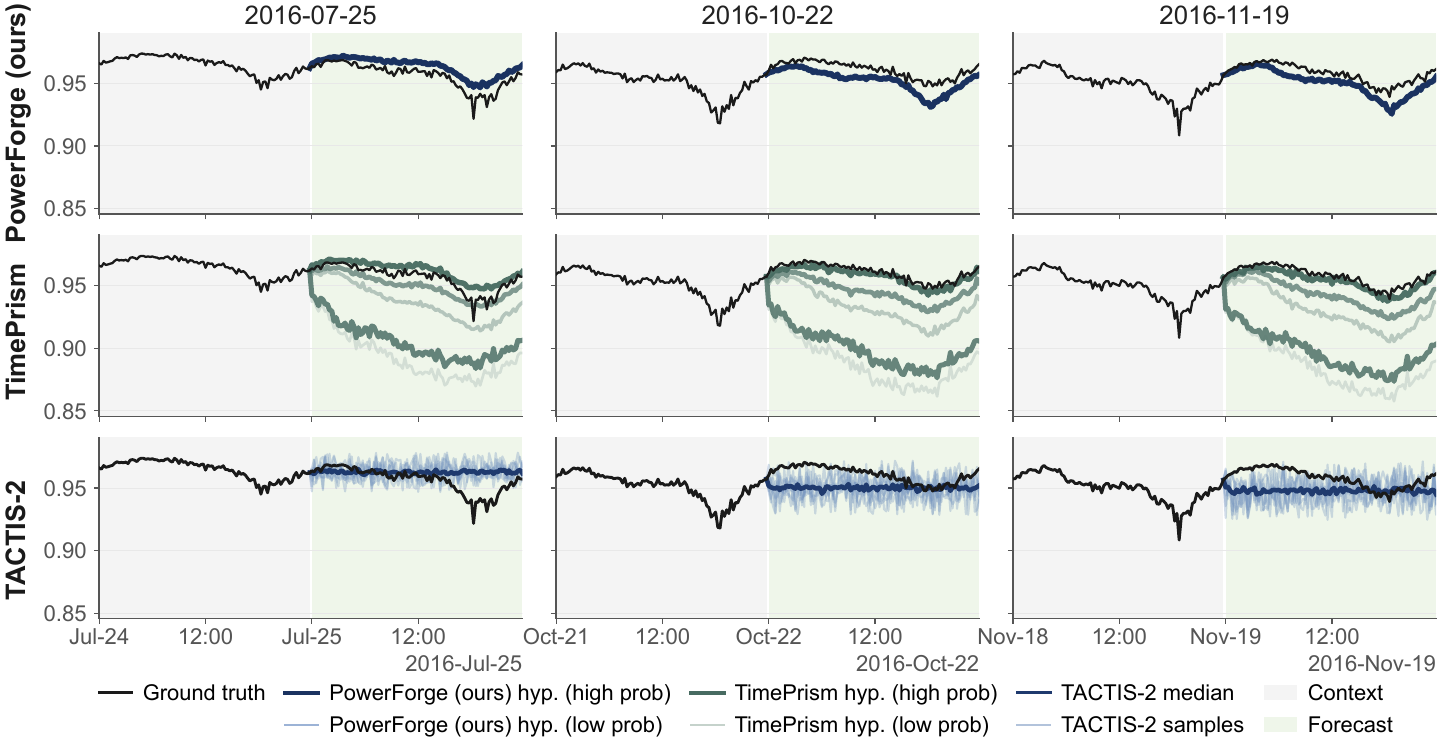}
    \caption{Voltage forecasts on Polish\,2383 ($9{,}532$ channels) across three test windows. PowerForge (top) produces a compact hypothesis fan tracking the diurnal pattern. TimePrism (middle) recovers the shape with wider spread. TACTiS-2 (bottom) flattens the daily cycle under per-step sample jitter.}
    \label{fig:qualitative}
\end{figure}
Density-based models behave inconsistently across scales. TACTiS-2 wins or ties on 500-bus security but its Safety\_mBrier rises from $0$ to $0.04$ at 3120-bus. TempFlow and Transformer-TempFlow show the inverse trend, weak on 500-bus security but reaching the second-best tier on 3120-bus. ETS, despite its simplicity, is surprisingly strong at small scale and degrades steadily with channel count. No baseline holds a stable position across both scales and axes, whereas PowerForge maintains a top-tier position throughout. At PEGASE\,9241 ($36{,}964$ channels), PowerForge is  still best on every metric (Appendix~\ref{app:9241}).

\subsection{Qualitative Analysis}
\label{sec:qualitative}

Figure~\ref{fig:qualitative} compares scenario forecasts from PowerForge, TimePrism, and TACTiS-2 on a voltage channel of Polish\,2383 across three test windows. PowerForge produces a compact set of hypotheses that track the diurnal shape across all windows, with probability mass aligned to the daily pattern. TimePrism recovers the overall shape but with a wider spread, and several low-probability hypotheses visibly undershoot the evening trough. Its winner-takes-all objective rewards hypothesis diversity, so the spread is partly by design. On this channel, the diversity translates into systematic underestimation rather than informative uncertainty. TACTiS-2 shows a different failure. Its median tracks the contextual mean reasonably well but flattens through the diurnal trough, while individual samples exhibit pronounced high-frequency jitter without coherent low-frequency motion, consistent with an autoregressive copula sampler that captures per-step marginals but attenuates temporal dependence over the 96-step horizon at this channel count. These behaviours align with the metric ordering in Table~\ref{tab:main}. Producing a structured set of trajectories that respects the low-frequency physics of the system remains the harder requirement for high-dimensional grid forecasting. Additional cross-grid trajectories and aggregate bus-level distributions are reported in Appendix~\ref{app:additional_safety}.

\subsection{Ablation Study}
\label{sec:ablation}

\begin{table}[h]
\centering
\small
\caption{Ablation on PEGASE\,1354 (single seed). $\Delta$ is the relative CRPS change against the full configuration, ordered by descending impact. CRPS and Distortion measure fidelity. Lower is better for both metrics. Full safety metrics are in Appendix~\ref{app:ablation_1354}.}
\label{tab:ablation}
\setlength{\tabcolsep}{8pt}
\begin{tabular}{lccc}
\toprule
Variant & CRPS $\downarrow$ & $\Delta$ & Distortion $\downarrow$ \\
\midrule
Full PowerForge & \textbf{0.0043} & -- & \textbf{0.0110} \\
$-$ anchor--delta parameterisation & 0.0080 & $+84\%$ & 0.0229 \\
$-$ quantile regularisation & 0.0059 & $+37\%$ & 0.0128 \\
$-$ ordered-quantile loss (use WTA) & 0.0055 & $+27\%$ & 0.0155 \\
$-$ physics regularisation & 0.0055 & $+27\%$ & 0.0120 \\
$-$ cross-type causal bridge & 0.0053 & $+22\%$ & 0.0120 \\
$-$ low-rank global token mixer & 0.0045 & $+4\%$ & 0.0154 \\
\bottomrule
\end{tabular}
\end{table}

We isolate the contribution of each PowerForge component by training
six ablated variants on PEGASE\,1354 with all remaining
hyperparameters fixed. Each variant disables a single architectural
element or training signal. Table~\ref{tab:ablation} reports CRPS
and Distortion together with the relative CRPS change against the
full configuration, ordered by descending impact. The anchor--delta voltage parameterisation is the single most important component. Removing it raises CRPS by $84\%$ and Distortion by more than $2\times$, confirming that operating in residual voltage space is essential when channels span the narrow physical band around $1.0$\,p.u. The next tier is calibration and training signal. The quantile-fan regularisation contributes $+37\%$ CRPS, replacing the ordered-quantile criterion with winner-takes-all training contributes $+27\%$, and removing physics regularisation contributes $+27\%$. The cross-type causal bridge has smaller direct impact at this scale ($+22\%$). The low-rank global mixer has the smallest CRPS impact ($+4\%$), but its effect on Distortion is substantial ($+40\%$, $0.0110 \to 0.0154$), indicating its role in single-best-hypothesis quality rather than overall distributional fit.


\section{Conclusion and Limitations}
\label{sec:conclusion}

\paragraph{Conclusion.} We introduced PowerPhase, a probabilistic forecasting
benchmark that brings transmission-scale multivariate systems
into reach of evaluation, and PowerForge, an
ordered-quantile scenario forecaster designed for this
regime. PowerPhase ships with physically typed channels,
AC-power-flow-simulated trajectories, and a voltage-risk-aware metric
suite that complements scoring rules.
PowerForge combines a reference-anchored residual
parameterisation, type-specific decoding heads coupled
through a causal cross-attention bridge, and a low-rank
global mixer that exchanges information across channels
through $K \ll C$ shared tokens at $O(CK)$ cost. Across the
five main grids and three seeds, it attains the best score on
19 of 20 grid--metric cells, ties with the strongest baseline
on the remaining one within seed variance, and ranks first on
average on every grid. The benchmark also surfaces a safety--fidelity trade-off:
distributional accuracy and constraint satisfaction rank
models differently, so no single average-case score is
sufficient for constraint-bound decisions.

\paragraph{Limitations.}
PowerPhase evaluates voltage-band risk rather than full AC feasibility. While target
trajectories are produced by AC power-flow solves, the proposed safety metrics only measure
violations of the $[0.95,1.05]$ p.u. voltage-magnitude band; they do not certify line limits,
generator reactive limits, power-balance residuals, angle consistency, or feasibility of each
predicted $(P,Q,V,\theta)$ scenario. The benchmark is also synthetic, driven by national
load and renewable traces, archetypal bus profiles, Gaussian perturbations, and fixed power
factors, so some safety--fidelity trends may reflect this simulator family. PowerForge encodes
variable type and a $P,Q\!\rightarrow\!V,\theta$ dependency, but not explicit topology or
admittance structure. Real-data validation, richer physical metrics, topology-aware models,
and broader large-scale baseline sweeps are left for future work.

%
%

\bibliographystyle{plainnat}
\bibliography{sample-base}

\newpage
\appendix
\phantomsection
\pdfbookmark[0]{Appendix}{appendix}
\section*{Appendix}
\titlecontents{section}[0em]
  {\vspace{0.5em}\bfseries} 
  {\thecontentslabel\quad}
  {}
  {\titlerule*[0.8pc]{.}\contentspage}

\titlecontents{subsection}[1.5em]
  {\normalfont\small}
  {\thecontentslabel\quad}
  {}
  {\titlerule*[0.8pc]{.}\contentspage}

\titlecontents{subsubsection}[3em]
  {\itshape\footnotesize}
  {\thecontentslabel\quad}
  {}
  {\titlerule*[0.8pc]{.}\contentspage}

\startcontents[appendix] 
\printcontents[appendix]{l}{1}{\setcounter{tocdepth}{3}}
\vspace{2em} 


\section{Benchmark Details}
\label{app:benchmark}

\subsection{Data Generation Pipeline Details}
\label{app:pipeline}

\subsubsection{Source Signals and Load Archetypes}
\label{app:source_archetypes}

All PowerPhase networks are driven by three national-level seed signals from the Open Power System Data archive \citep{wiese2019open}: aggregate German load (\texttt{DE\_load\_actual\_entsoe\_transparency}), solar generation (\texttt{DE\_solar\_generation\_actual}), and wind generation (\texttt{DE\_wind\_generation\_actual}). We retain 2015--2016 at $15$-minute resolution and fill missing entries by linear interpolation followed by forward and backward fill, yielding $70{,}176$ time steps per signal.

Each load bus is classified as \emph{high-voltage} (HV) if its nominal voltage exceeds $110\,\mathrm{kV}$ or its nominal active power exceeds $10\,\mathrm{MW}$, and as \emph{low-voltage} (LV) otherwise. Five normalised daily-shape profiles are constructed from the seed signals:
\begin{enumerate}[leftmargin=1.5em,itemsep=2pt]
    \item \textbf{Baseline}: the national load curve, normalised to unit mean.
    \item \textbf{Industrial}: a flattened version of the national load, $\mu + \alpha_{\mathrm{ind}}(L_t - \mu)$ with $\alpha_{\mathrm{ind}}=0.3$, normalised to unit mean.
    \item \textbf{PV-correlated}: national load minus $k_{\mathrm{pv}} \cdot \text{solar}$ with $k_{\mathrm{pv}}=3.0$, normalised to unit mean.
    \item \textbf{Wind-correlated}: national load minus $k_{\mathrm{wind}} \cdot \text{wind}$ with $k_{\mathrm{wind}}=2.0$, normalised to unit mean.
    \item \textbf{EV-correlated}: the baseline profile boosted by $+40\%$ during $18$:$00$--$22$:$00$, normalised to unit mean.
\end{enumerate}
Each HV bus is assigned the Industrial profile with probability $0.8$ and the Baseline profile with probability $0.2$. Each LV bus draws from $\{$Baseline, PV, EV, Wind$\}$ with base weights $(0.4, 0.3, 0.2, 0.1)$, modulated by a deterministic region index $r = \texttt{bus\_id} \bmod 3$: region~$0$ doubles the PV weight and region~$1$ doubles the Wind weight, with weights renormalised before sampling.

\subsubsection{Noise Model}
\label{app:noise}

The per-bus active power injection at time step $t$ is
\begin{equation}
    P_{n,t} = P_n^{\mathrm{nom}} \cdot s_{n,t} \cdot (1 + \varepsilon^{\mathrm{reg}}_{r(n),t} + \varepsilon^{\mathrm{node}}_{n,t}),
    \label{eq:injection}
\end{equation}
where $s_{n,t}$ is the normalised archetype shape, $\varepsilon^{\mathrm{reg}}_{r,t} \sim \mathcal{N}(0, \sigma_{\mathrm{reg}})$ is a spatially correlated regional noise term shared across all buses with region index $r(n)$, and $\varepsilon^{\mathrm{node}}_{n,t} \sim \mathcal{N}(0, \sigma_n)$ is independent per-node noise. Negative injections are clipped to zero. Standard deviations are $\sigma_{\mathrm{reg}} = 0.03$, $\sigma_n = 0.02$ for HV nodes, and $\sigma_n = 0.08$ for LV nodes.

\subsubsection{Reactive Power and AC Power-Flow Solve}
\label{app:solver}

Reactive power is derived from active power via a per-bus power factor $\cos\phi_n$ sampled uniformly at initialisation and held fixed across all time steps,
\begin{equation}
    Q_{n,t} = P_{n,t} \cdot \tan(\arccos(\cos\phi_n)),
\end{equation}
with $\cos\phi_n \in [0.96, 0.99]$ for HV buses and $[0.90, 0.99]$ for LV buses.

For every time step we run AC power flow through pandapower \citep{thurner2018pandapower} using Newton--Raphson. A magnitude back-off and an Iwamoto multiplier fallback are applied when Newton--Raphson fails to converge, and any remaining failed steps are forward-filled with the last converged state. Convergence rates exceed $99\%$ on every PowerPhase network. Several PEGASE test cases contain generators with numerically zero $Q_{\max} - Q_{\min}$ ranges, which we widen to $5000\,\mathrm{MVAr}$ before running the pipeline to avoid division-by-zero in the solver.

\subsection{Normalisation Conventions}
\label{app:normalisation}

The four variable types are normalised independently before computing all evaluation metrics:

\begin{center}
\small
\begin{tabular}{lll}
\toprule
Variable & Unit & Normalisation \\
\midrule
$P$ (active power)    & MW   & divided by training-set $\max|P|$ \\
$Q$ (reactive power)  & MVAr & divided by training-set $\max|Q|$ \\
$V$ (voltage magnitude) & p.u. & kept as-is \\
$\theta$ (voltage angle) & rad  & kept as-is (converted from degrees) \\
\bottomrule
\end{tabular}
\end{center}

The max-abs constants for $P$ and $Q$ are computed on the training split only and reused on validation and test, so no information from the evaluation period leaks into normalisation.

\subsection{Metric Definitions}
\label{app:metrics}

Throughout this section, let $Z \in \mathbb{R}^{B \times T \times D}$ be the ground-truth tensor over $B$ prediction windows, $T$ forecast steps, and $D$ channels. A forecaster produces $K$ scenarios $\hat Z^{(k)}$ with weights $w^{(k)}$ satisfying $\sum_k w^{(k)}_{b,d} = 1$. For sample-based methods, $w^{(k)} = 1/K$ uniformly.

\subsubsection{CRPS}
\label{app:crps}

We compute CRPS via the energy-score representation,
\begin{equation}
    \mathrm{CRPS}_{b,t,d} = \sum_k w^{(k)}_{b,d} \bigl|\hat Z^{(k)}_{b,t,d} - Z_{b,t,d}\bigr| - \tfrac{1}{2}\,\mathbb{E}_w\bigl[|\hat Z - \hat Z'|\bigr],
    \label{eq:crps_energy}
\end{equation}
and exploit the sorted-CDF identity for efficient computation of the second term. With scenarios sorted in ascending order $\hat Z^{(\pi_1)} \le \hat Z^{(\pi_2)} \le \cdots$ and cumulative weights $F_i = \sum_{j=1}^{i} w^{(\pi_j)}_{b,d}$,
\begin{equation}
    \mathbb{E}_w\bigl[|\hat Z - \hat Z'|\bigr] = 2 \sum_{i=1}^{K-1} F_i (1 - F_i) \bigl(\hat Z^{(\pi_{i+1})}_{b,t,d} - \hat Z^{(\pi_i)}_{b,t,d}\bigr),
\end{equation}
which avoids the $K \times K$ pairwise difference matrix. The reported CRPS averages $\mathrm{CRPS}_{b,t,d}$ over all windows, steps, and channels.

\subsubsection{Distortion}
\label{app:distortion}

Distortion measures the quality of the single best hypothesis in each prediction window:
\begin{equation}
    \mathrm{Distortion} = \frac{1}{B} \sum_{b=1}^{B} \min_{k} \sqrt{\frac{1}{T D} \sum_{t,d} \bigl(\hat Z^{(k)}_{b,t,d} - Z_{b,t,d}\bigr)^2}.
\end{equation}
This is the standard winner-takes-all measure used in multiple-choice learning \citep{cortes2025winnertakesall,lee2016stochastic}.

\subsubsection{Voltage Safety Metrics}
\label{app:safety_metrics}

All safety metrics are computed over voltage channels, indexed as the third coordinate of the interleaved $[P, Q, V, \theta]$ layout. Let $\mathcal{V}$ index $(b, t, d)$ triples on voltage channels, and $[V_{\min}, V_{\max}] = [0.95, 1.05]\,\mathrm{p.u.}$. The ground-truth violation event is $Y_{b,t,d} = \mathbf{1}\{V_{b,t,d} \notin [V_{\min}, V_{\max}]\}$ and the per-scenario indicator is $\hat Y^{(k)}_{b,t,d} = \mathbf{1}\{\hat V^{(k)}_{b,t,d} \notin [V_{\min}, V_{\max}]\}$. The per-scenario violation magnitude is
\begin{equation}
    \delta^{(k)}_{b,t,d} = \max(0, V_{\min} - \hat V^{(k)}_{b,t,d}) + \max(0, \hat V^{(k)}_{b,t,d} - V_{\max}).
\end{equation}

The three voltage safety metrics are then
\begin{align}
    \mathrm{Safety\_mBrier} &= \frac{1}{|\mathcal{V}|} \sum_{(b,t,d) \in \mathcal{V}} \sum_k w^{(k)}_{b,d} \bigl(\hat Y^{(k)}_{b,t,d} - Y_{b,t,d}\bigr)^2, \label{eq:safety_mbrier}\\
    \mathrm{NECV} &= \frac{1}{|\mathcal{V}|} \sum_{(b,t,d) \in \mathcal{V}} \sum_k w^{(k)}_{b,d} \, \delta^{(k)}_{b,t,d}, \label{eq:necv}\\
    \mathrm{CVaR}_{0.1} &= \frac{1}{|\mathcal{V}|} \sum_{(b,t,d) \in \mathcal{V}} \frac{1}{\lceil 0.1 K \rceil} \sum_{k \in \mathcal{T}_{0.1}(b,t,d)} \delta^{(k)}_{b,t,d}, \label{eq:cvar}
\end{align}
where $\mathcal{T}_{0.1}(b, t, d)$ selects the $\lceil 0.1 K \rceil$ scenarios with the largest $\delta^{(k)}$ at $(b, t, d)$. The tail average for $\mathrm{CVaR}_{0.1}$ uses uniform weights, matching the discrete CVaR estimator of \citep{rockafellar2000cvar}; weighting the tail by $w^{(k)}$ would suppress low-probability extreme scenarios and contradict the risk-averse intent.

\paragraph{Estimator choice for Safety\_mBrier.} The per-scenario form in Eq.~\eqref{eq:safety_mbrier} differs from the standard Brier on the weighted violation probability $\bar Y_{b,t,d} = \sum_k w^{(k)}_{b,d} \hat Y^{(k)}_{b,t,d}$ by a non-negative scenario-disagreement term, and reduces to the standard Brier in the deterministic limit. We retain the per-scenario form because grid contingency screening consumes scenarios individually rather than as a single aggregate violation probability. Each scenario is matched to a concrete remedial-action plan, so the operationally relevant quantity is the average forecast quality across scenarios as they are screened, not the calibration of the aggregate. A forecasting model whose mean violation probability is calibrated but whose scenarios disagree is operationally worse than one whose scenarios agree at the same mean, because the disagreement leaves the operator without a clear contingency to act on. The estimator is also unbiased on $K$ weighted scenarios.

\paragraph{Interpretation.}
Safety\_mBrier assesses \emph{detection} (does the forecaster know when violations occur), $\mathrm{NECV}$ assesses \emph{average severity}, and $\mathrm{CVaR}_{0.1}$ assesses \emph{tail severity}. A forecaster can score well on CRPS while scoring poorly on these metrics if it concentrates probability mass near the conditional mean and underestimates the tails of the voltage distribution.

\subsection{Bus Type Convention}
\label{app:bus_types}

Each bus is classified into one of three structural types:

\begin{center}
\small
\begin{tabular}{clll}
\toprule
Type & Name & Known (input) & Solved (output) \\
\midrule
1 & Slack (reference) & $V$, $\theta$ & $P$, $Q$ \\
2 & PV (generator)    & $P$, $V$      & $Q$, $\theta$ \\
3 & PQ (load)         & $P$, $Q$      & $V$, $\theta$ \\
\bottomrule
\end{tabular}
\end{center}

For PQ buses (the majority in all PowerPhase networks), $P$ and $Q$ are the exogenous inputs to the power-flow solver and $V$, $\theta$ are the physical responses. This structural asymmetry motivates the causal $P, Q \to V, \theta$ bridge in PowerForge (\S\ref{sec:powerforge}).

\subsection{Train / Validation / Test Splits}
\label{app:splits}

All six PowerPhase networks share the same temporal split applied to the $70{,}176$-step record (2015-01-01 to 2016-12-31 at 15-minute resolution):

\begin{center}
\small
\begin{tabular}{lll}
\toprule
Split & Period & Purpose \\
\midrule
Train      & 2015-01-01 to 2016-06-30 & Model training \\
Validation & Last 10 windows of train & Checkpoint selection \\
Test       & 2016-07-01 to 2016-12-31 & Final evaluation \\
\bottomrule
\end{tabular}
\end{center}

Test evaluation uses rolling-origin testing with $10$ prediction windows of length $T_p = 96$ (one day). Each window uses a context of $T_h = 672$ steps (seven days) immediately preceding the prediction horizon. All reported metrics are averaged over these $10$ windows and three random seeds.

\section{PowerForge Architecture}
\label{app:architecture}

\subsection{Anchor-Delta Details}
\label{app:anchor_delta}

The reference $Z^{\mathrm{ref}}$ in Eq.~(\ref{eq:anchor}) is computed from the last $S \times L$ input steps reshaped into $S$ non-overlapping daily segments of length $L = 96$, with $S = 7$ covering one week. The query $X^{\mathrm{query}}_c$ is the last $Q = 8$ steps of the input, and the key for segment $s$ is the last $Q$ steps of that segment. The Pearson correlation $\rho$ between query and key is computed per channel with an $\epsilon = 10^{-8}$ stabiliser on the variance, and the temperature in the softmax is $\tau = 0.5$.

The reference day is tiled to match the input and prediction lengths, giving $Z^{\mathrm{ref}}_{\mathrm{input}}$ and $Z^{\mathrm{ref}}_{\mathrm{future}}$. The model consumes $x - Z^{\mathrm{ref}}_{\mathrm{input}}$ as input and emits residuals $\hat\delta$, recovered as $\hat y = \hat\delta + Z^{\mathrm{ref}}_{\mathrm{future}}$. For $\theta$ channels, both bias-correction differences and recovered values are wrapped to $[-\pi, \pi]$ via $\operatorname{atan2}(\sin, \cos)$ to handle angular discontinuity.

A bias-correction term, the mean residual $x - Z^{\mathrm{ref}}_{\mathrm{input}}$ over the last $W = 8$ input steps, is added to both $Z^{\mathrm{ref}}_{\mathrm{input}}$ and $Z^{\mathrm{ref}}_{\mathrm{future}}$ before forming the residual. This absorbs slow drift between the periodic reference and the most recent observations. For $\theta$ channels, all reference and residual differences in this subsection are wrapped via $\operatorname{atan2}(\sin, \cos)$, with full details in Appendix~\ref{app:heads}.

\subsection{Type-Specific Heads}
\label{app:heads}

Each per-type head consumes the fused state $h^{(m)}_c = h_c + e_m$ and emits a residual prediction $\delta^{(m)}_c$ over the $T_p$-step horizon.

\paragraph{\texorpdfstring{$P$}{P} and \texorpdfstring{$Q$}{Q} heads.}
The $P$ and $Q$ heads are unbounded and parameterised as
\begin{equation}
\delta^{(m)}_c = \mu^{(m)}_c + \sigma^{(m)}_c \, \epsilon^{(m)},
\label{eq:pq_head}
\end{equation}
where $\mu^{(m)}_c$ is a linear projection of the fused state, $\sigma^{(m)}_c$ is a softplus-activated projection, and $\epsilon^{(m)}$ is a per-scenario noise vector. The default distribution is $\epsilon^{(m)} \sim \mathcal{N}(0, 1)$ resampled per batch, with a Student-$t$ alternative ($\nu = 3$) also supported. The noise term acts as a stochastic regulariser rather than the source of scenario diversity. Diversity across the $m$ axis is driven by the $M$ learnable scenario embeddings $\{e_m\}$, which differentiate the branches at the input of the head, and is preserved by the pointwise sorting step in Section~\ref{sec:powerforge_training} that assigns each branch to a quantile level.

\paragraph{\texorpdfstring{$V$}{V} and \texorpdfstring{$\theta$}{theta} heads.}
The $V$ head passes a learned projection $v^{(m)}_c$ through a $\tanh$ gate scaled by a learnable per-type magnitude $\Delta_V$,
\begin{equation}
\delta^{(m)}_{V, c} = \Delta_V \cdot \tanh\bigl(v^{(m)}_c\bigr),
\end{equation}
with $\Delta_V$ initialised at $0.05$ and floored at $10^{-4}$. The initialisation reflects the operationally narrow voltage band around $1.0$\,p.u., and the floor prevents the gate from collapsing to zero during training. The $\theta$ head uses the same tanh-gate form with a fixed $\Delta_\theta = \pi$. Both gates structurally bound the residual without an explicit penalty.

\paragraph{Angular wrapping.}
For $\theta$ channels, both the bias-correction differences in Appendix~\ref{app:anchor_delta} and the recovered absolute predictions $\hat y_\theta = \hat\delta_\theta + Z^{\mathrm{ref}}_{\mathrm{future}, \theta}$ are wrapped to $[-\pi, \pi]$ via $\operatorname{atan2}(\sin, \cos)$ to handle the discontinuity at $\pm\pi$.

\subsection{Channel Encoder}
\label{app:encoder}

The channel encoder maps the input history $x \in \mathbb{R}^{B \times T \times C}$ into a per-channel representation $h \in \mathbb{R}^{B \times C \times H}$ through three additive sources of information.

\paragraph{Multi-scale temporal backbone.}
Each channel is processed independently by a shared temporal encoder. The raw signal and three 1-D convolutional filters with kernel sizes $k \in \{5, 25, 97\}$ extract features at $1.25$-hour, $6.25$-hour, and $1$-day receptive fields at $15$-minute resolution. Their outputs are concatenated along the feature axis and projected to dimension $H$ through a two-layer GELU MLP. A per-channel scale token, obtained by passing the mean absolute value of each channel through a small MLP, is added to provide scale awareness across heterogeneous variable types. The result is layer-normalised with dropout $0.1$.

\paragraph{Node, variable, and calendar features.}
Each channel additionally carries a learned node embedding ($d_{\text{node}} = 8$) and a variable-type embedding ($d_{\text{var}} = 2$). These two embeddings are concatenated and projected to dimension $H$, then added to the backbone output. They provide positional identity (which bus) and type identity (which of $P, Q, V, \theta$). Calendar features (time of day, day of week) are averaged over the input window, projected to dimension $H$, and added as a shared offset across channels.

\subsection{Training Objective Details}
\label{app:training}

\subsubsection{Full Loss Formulation}
\label{app:full_loss}
The total training loss is
\begin{equation}
    \mathcal{L} = \mathcal{L}_{\mathrm{OQ}} + \lambda_V\,\mathcal{L}_{\mathrm{phys},V} + \lambda_\theta\,\mathcal{L}_{\mathrm{phys},\theta} + w_{\mathrm{width}}\,\mathcal{L}_{\mathrm{width}} + w_{\mathrm{jitter}}\,\mathcal{L}_{\mathrm{jitter}} + \mathcal{L}_{\mathrm{adapt}} + \mathcal{L}_{\mathrm{Beta}}+ w_{\mathrm{gate}}\,\mathcal{L}_{\mathrm{gate}}.
    \label{eq:total_loss}
\end{equation}
The Beta-prior term $\mathcal{L}_{\mathrm{Beta}}$ keeps the learnable Beta$(\alpha, \beta)$ that parameterises $\tau_m$ close to its initial mean and scale, with weights $5 \times 10^{-3}$ and $10^{-3}$ respectively. The gate term $\mathcal{L}_{\mathrm{gate}}$ is an $\ell_2$ penalty on the V/$\theta$ pre-backbone context gates toward their initialisation, with $w_{\mathrm{gate}} = 2\times10^{-4}$.

\subsubsection{Ordered-Quantile Loss \texorpdfstring{($\mathcal{L}_{\mathrm{OQ}}$)}{(LOQ)}}

\label{app:oq_loss}
Given $M$ hypotheses and quantile levels $\tau_1 < \cdots < \tau_M$ adapted during training through a learnable Beta-base law, the hypotheses are sorted along the scenario axis per channel and a pinball loss is applied:
\begin{equation}
    \mathcal{L}_{\mathrm{OQ}} = \frac{1}{B M T_p C} \sum_{b, m, t, c} \rho_{\tau_m}\!\bigl(Z_{b, t, c} - \hat{Z}^{(m)}_{b, t, c}\bigr),
\end{equation}
where $\rho_\tau(u) = u(\tau - \mathbf{1}\{u < 0\})$ and $\hat{Z}^{(m)}$ is the $m$-th sorted hypothesis.

\subsubsection{Physics Regularisation \texorpdfstring{($\mathcal{L}_{\mathrm{phys}}$)}{(Lphys)}}
\label{app:phys_loss}
$\mathcal{L}_{\mathrm{phys},V}$ applies a smooth-$L_1$ penalty toward zero on the voltage residual $\delta^{(m)}_{c,t}$ normalised by the learnable scale $\Delta_V$ (detached during this computation). This encourages voltage residuals to stay small relative to the gate width. $\mathcal{L}_{\mathrm{phys},\theta}$ penalises the wrapped temporal difference $|\mathrm{wrap}_{[-\pi, \pi]}(\hat\theta^{(m)}_{c, t+1} - \hat\theta^{(m)}_{c, t})|$, which prevents $\pm\pi$ jumps from being scored as oscillation. Both are averaged over scenarios, time, and channels of the relevant variable type. Default weights are $\lambda_V = 0.05$ and $\lambda_\theta = 0.01$.

\subsubsection{Auxiliary Regularisers}
\label{app:aux_reg}
\paragraph{Quantile width.}
$\mathcal{L}_{\mathrm{width}}$ caps the inter-quantile spread between two reference quantile levels (default $\tau_{\mathrm{low}} = 0.1$ and $\tau_{\mathrm{high}} = 0.9$, taken as the closest hypotheses on the learned grid) at $\gamma$ times a per-window per-channel target standard deviation $\sigma_{b,c}$:
\begin{equation}
    \mathcal{L}_{\mathrm{width}} = \frac{1}{B T_p C} \sum_{b, t, c} \mathrm{ReLU}\!\Bigl(\frac{|\hat{Z}^{(\mathrm{high})}_{b,t,c} - \hat{Z}^{(\mathrm{low})}_{b,t,c}|}{\gamma\,\sigma_{b,c}} - 1\Bigr),
\end{equation}
with $\gamma = 1.0$ and $w_{\mathrm{width}} = 0.002$. The relative form prevents per-channel scale heterogeneity from biasing the penalty.
\paragraph{Temporal jitter.}
$\mathcal{L}_{\mathrm{jitter}}$ discourages step-to-step oscillation of the median-centred quantile fan. After subtracting the median quantile, the temporal differences of the remaining quantiles are normalised by a per-window per-channel target standard deviation and averaged. The median itself is masked from the average. Default weight is $w_{\mathrm{jitter}} = 0.001$.
\paragraph{Adaptive variable weights.}
A learnable vector $\omega \in \mathbb{R}^4$ scales the loss contribution of each variable type. It is initialised from the inverse per-type volatility on the first $40$ training batches and clamped to $[0.3, 3.0]$. An $\ell_2$ penalty with weight $w_{\mathrm{adapt}} = 2 \times 10^{-4}$ regularises $\omega$ toward unity.

\section{Training and Implementation}
\label{app:training_impl}

\subsection{Hyperparameters}
\label{app:hyperparameters}

Table~\ref{tab:hyperparams} lists the main hyperparameters used across all PowerPhase networks.

\begin{table}[h]
    \centering
    \small
    \caption{PowerForge hyperparameters (shared across all six PowerPhase networks unless noted otherwise).}
    \label{tab:hyperparams}
    \begin{tabular}{llr}
    \toprule
    Group & Parameter & Value \\
    \midrule
    \multirow{4}{*}{Architecture}
    & Hidden dimension $H$ & $128$ \\
    & Backbone kernel sizes & $\{5, 25, 97\}$ \\
    & Segment length & $96$ \\
    & Global tokens $K$ & $64$ \\
    \midrule
    \multirow{3}{*}{Embeddings}
    & Node embedding dim & $8$ \\
    & Variable embedding dim & $2$ \\
    & Backbone dropout & $0.1$ \\
    \midrule
    \multirow{3}{*}{Decoder}
    & Number of hypotheses $M$ & $16$ \\
    & Causal bridge scale init & $1.0$ \\
    & Scenario scoring & linear head \\
    \midrule
    \multirow{3}{*}{Anchor}
    & Number of segments $S$ & $7$ \\
    & Attention query length & $8$ \\
    & Attention temperature $\tau$ & $0.5$ \\
    \midrule
    \multirow{4}{*}{Training}
    & Objective mode & ordered-quantile \\
    & Quantile levels & linspace($0.05, 0.95, 16$) \\
    & Learning rate & $10^{-3}$ \\
    & Weight decay & $10^{-6}$ \\
    \midrule
    \multirow{5}{*}{Loss weights}
    & Physics voltage $\lambda_V$ & $0.05$ \\
    & Physics angle $\lambda_\theta$ & $0.01$ \\
    & Width regulariser & $0.002$ \\
    & Jitter regulariser & $0.001$ \\
    & Adaptive weight reg & $2 \times 10^{-4}$ \\
    \midrule
    \multirow{2}{*}{V/$\theta$ gate}
    & Voltage $\Delta_V$ init & $0.05$ \\
    & Angle scale & $\pi$ (fixed) \\
    \midrule
    \multirow{2}{*}{Input}
    & Context length & $672$ steps \\
    & Use full history & True \\
    \bottomrule
    \end{tabular}
\end{table}

\subsection{Implementation Details}
\label{app:implementation}
\begin{table}[h]
    \centering
    \small
    \caption{Per-network training batch size for PowerForge.}
    \label{tab:batch_size}
    \begin{tabular}{lr}
    \toprule
    Network & Batch size \\
    \midrule
    ACTIVSg\,500 & $32$ \\
    PEGASE\,1354 & $32$ \\
    Polish\,2383 & $16$ \\
    PEGASE\,2869 & $32$ \\
    Polish\,3120 & $32$ \\
    PEGASE\,9241 & $8$ \\
    \bottomrule
    \end{tabular}
\end{table}
\subsubsection{Per-Network Batch Size}
\label{app:batch}

Batch sizes are chosen empirically per network to fit within the $96$\,GB GPU memory of the NVIDIA RTX PRO 6000 Blackwell GPU. Larger networks use smaller batches due to the linear scaling of activation memory with channel count. Table~\ref{tab:batch_size} reports the per-network values.

\subsubsection{Train, Validation, and Test Splits}
\label{app:splits}

All six PowerPhase networks share the same temporal split applied to the $70{,}176$-step record (2015-01-01 to 2016-12-31 at 15-minute resolution). The first $18$ months form the training set, the last $6$ months form the test set, and the final $10$ windows of the training period are reserved for validation. Test evaluation uses rolling-origin testing with $10$ equally spaced prediction windows of length $T_p = 96$ (one day). Each window uses a context of $T_h = 672$ steps (seven days) immediately preceding the prediction horizon. All reported metrics are averaged over these $10$ windows and three random seeds.

\begin{table*}[h]
\caption{Probabilistic forecasting results on \textsc{PowerPhase} across five grids. CRPS, Distortion, and MSE measure \emph{fidelity}; Safety\_mBrier, NECV, and $\text{CVaR}_{0.1}$ measure \emph{safety}. All values are mean{\scriptsize$\pm$std} over three seeds. \textbf{Bold}: best; \underline{underline}: second best.}
\label{tab:powerphase_five_grids}
\centering
\setlength{\tabcolsep}{2.3pt}
\scriptsize
\resizebox{\textwidth}{!}{%
\begin{tabular}{@{}l l c c c c c c@{}}
\toprule
 & & \multicolumn{3}{c}{\emph{Fidelity}\,$\downarrow$}
   & \multicolumn{3}{c}{\emph{Safety}\,$\downarrow$} \\
\cmidrule(lr){3-5} \cmidrule(lr){6-8}
\textbf{Grid} & \textbf{Model}
  & CRPS & Distortion & MSE
  & Safety\_mBrier & NECV & $\text{CVaR}_{0.1}$ \\
\midrule
  500-bus & DeepAR
    & 0.1156{\scriptsize$\pm$0.0240}
    & 0.5537{\scriptsize$\pm$0.1310}
    & 0.0049{\scriptsize$\pm$0.0013}
    & 0.9519{\scriptsize$\pm$0.0233}
    & 0.8220{\scriptsize$\pm$0.2702}
    & 2.1753{\scriptsize$\pm$0.6901} \\

   & ETS
    & 0.0064{\scriptsize$\pm$0.0000}
    & 0.0481{\scriptsize$\pm$0.0000}
    & 0.0006{\scriptsize$\pm$0.0000}
    & 0.0561{\scriptsize$\pm$0.0002}
    & 0.0026{\scriptsize$\pm$0.0000}
    & \underline{0.0116}{\scriptsize$\pm$0.0000} \\

   & TimeMCL
    & \underline{0.0054}{\scriptsize$\pm$0.0003}
    & \underline{0.0118}{\scriptsize$\pm$0.0001}
    & 0.0003{\scriptsize$\pm$0.0002}
    & \underline{0.0200}{\scriptsize$\pm$0.0200}
    & 0.0191{\scriptsize$\pm$0.0191}
    & 0.1906{\scriptsize$\pm$0.1907} \\

   & TimePrism
    & 0.0077{\scriptsize$\pm$0.0003}
    & 0.0195{\scriptsize$\pm$0.0023}
    & 0.0004{\scriptsize$\pm$0.0002}
    & 0.0221{\scriptsize$\pm$0.0062}
    & \underline{0.0014}{\scriptsize$\pm$0.0008}
    & 2.9950{\scriptsize$\pm$1.1445} \\

   & TACTiS-2
    & 0.0057{\scriptsize$\pm$0.0002}
    & 0.0139{\scriptsize$\pm$0.0006}
    & \underline{0.0002}{\scriptsize$\pm$0.0000}
    & \textbf{0.0000}{\scriptsize$\pm$0.0000}
    & \textbf{0.0000}{\scriptsize$\pm$0.0000}
    & \textbf{0.0000}{\scriptsize$\pm$0.0000} \\

   & TempFlow
    & 0.0144{\scriptsize$\pm$0.0008}
    & 0.0679{\scriptsize$\pm$0.0005}
    & 0.0004{\scriptsize$\pm$0.0001}
    & 0.2740{\scriptsize$\pm$0.0519}
    & 0.0075{\scriptsize$\pm$0.0018}
    & 0.0332{\scriptsize$\pm$0.0043} \\

   & Trans-TempFlow
    & 0.0146{\scriptsize$\pm$0.0006}
    & 0.0702{\scriptsize$\pm$0.0011}
    & 0.0004{\scriptsize$\pm$0.0001}
    & 0.2896{\scriptsize$\pm$0.0581}
    & 0.0081{\scriptsize$\pm$0.0024}
    & 0.0380{\scriptsize$\pm$0.0100} \\

   & TimeGrad
    & 0.0160{\scriptsize$\pm$0.0011}
    & 0.0954{\scriptsize$\pm$0.0092}
    & 0.0010{\scriptsize$\pm$0.0002}
    & 0.7630{\scriptsize$\pm$0.0192}
    & 0.0977{\scriptsize$\pm$0.0122}
    & 0.3316{\scriptsize$\pm$0.0358} \\

   & \textbf{PowerForge (Ours)}
    & \textbf{0.0030}{\scriptsize$\pm$0.0008}
    & \textbf{0.0072}{\scriptsize$\pm$0.0019}
    & \textbf{0.0001}{\scriptsize$\pm$0.0000}
    & \textbf{0.0000}{\scriptsize$\pm$0.0000}
    & \textbf{0.0000}{\scriptsize$\pm$0.0000}
    & \textbf{0.0000}{\scriptsize$\pm$0.0000} \\
\midrule
  1354-bus & DeepAR
    & 0.1214{\scriptsize$\pm$0.0342}
    & 0.6547{\scriptsize$\pm$0.1231}
    & 0.0071{\scriptsize$\pm$0.0013}
    & 0.8139{\scriptsize$\pm$0.0766}
    & 0.3411{\scriptsize$\pm$0.2354}
    & 0.9371{\scriptsize$\pm$0.5997} \\

   & ETS
    & 0.0088{\scriptsize$\pm$0.0000}
    & 0.0500{\scriptsize$\pm$0.0000}
    & 0.0007{\scriptsize$\pm$0.0000}
    & 0.1506{\scriptsize$\pm$0.0007}
    & 0.0069{\scriptsize$\pm$0.0000}
    & 0.0292{\scriptsize$\pm$0.0000} \\

   & TimeMCL
    & 0.0097{\scriptsize$\pm$0.0010}
    & \underline{0.0209}{\scriptsize$\pm$0.0002}
    & 0.0005{\scriptsize$\pm$0.0000}
    & 0.0237{\scriptsize$\pm$0.0123}
    & 0.0073{\scriptsize$\pm$0.0055}
    & 0.0642{\scriptsize$\pm$0.0548} \\

   & TimePrism
    & 0.0103{\scriptsize$\pm$0.0017}
    & 0.0216{\scriptsize$\pm$0.0009}
    & 0.0026{\scriptsize$\pm$0.0023}
    & 0.0403{\scriptsize$\pm$0.0254}
    & 0.0270{\scriptsize$\pm$0.0316}
    & 5.0594{\scriptsize$\pm$2.9581} \\

   & TACTiS-2
    & \underline{0.0059}{\scriptsize$\pm$0.0001}
    & 0.0251{\scriptsize$\pm$0.0010}
    & \underline{0.0004}{\scriptsize$\pm$0.0000}
    & \underline{0.0118}{\scriptsize$\pm$0.0006}
    & \underline{0.0008}{\scriptsize$\pm$0.0000}
    & \underline{0.0011}{\scriptsize$\pm$0.0000} \\

   & TempFlow
    & 0.0148{\scriptsize$\pm$0.0073}
    & 0.0571{\scriptsize$\pm$0.0274}
    & 0.0009{\scriptsize$\pm$0.0004}
    & 0.2822{\scriptsize$\pm$0.2348}
    & 0.0108{\scriptsize$\pm$0.0086}
    & 0.0290{\scriptsize$\pm$0.0242} \\

   & Trans-TempFlow
    & 0.0206{\scriptsize$\pm$0.0007}
    & 0.0743{\scriptsize$\pm$0.0003}
    & 0.0013{\scriptsize$\pm$0.0001}
    & 0.4693{\scriptsize$\pm$0.0352}
    & 0.0197{\scriptsize$\pm$0.0027}
    & 0.0464{\scriptsize$\pm$0.0042} \\

   & TimeGrad
    & 0.0109{\scriptsize$\pm$0.0048}
    & 0.0945{\scriptsize$\pm$0.0575}
    & 0.0023{\scriptsize$\pm$0.0024}
    & 0.2496{\scriptsize$\pm$0.1912}
    & 0.0493{\scriptsize$\pm$0.0471}
    & 0.3581{\scriptsize$\pm$0.2910} \\

   & \textbf{PowerForge (Ours)}
    & \textbf{0.0040}{\scriptsize$\pm$0.0003}
    & \textbf{0.0106}{\scriptsize$\pm$0.0003}
    & \textbf{0.0001}{\scriptsize$\pm$0.0000}
    & \textbf{0.0030}{\scriptsize$\pm$0.0000}
    & \textbf{0.0007}{\scriptsize$\pm$0.0000}
    & \textbf{0.0008}{\scriptsize$\pm$0.0000} \\
\midrule
  2383-bus & DeepAR
    & 0.0583{\scriptsize$\pm$0.0261}
    & 0.4124{\scriptsize$\pm$0.1674}
    & 0.0037{\scriptsize$\pm$0.0038}
    & 0.5440{\scriptsize$\pm$0.1632}
    & 0.0535{\scriptsize$\pm$0.0465}
    & 0.1831{\scriptsize$\pm$0.1303} \\

   & ETS
    & 0.0111{\scriptsize$\pm$0.0000}
    & 0.1513{\scriptsize$\pm$0.0000}
    & 0.0058{\scriptsize$\pm$0.0000}
    & 0.0352{\scriptsize$\pm$0.0001}
    & 0.0029{\scriptsize$\pm$0.0000}
    & 0.0099{\scriptsize$\pm$0.0000} \\

   & TimeMCL
    & 0.0070{\scriptsize$\pm$0.0008}
    & \underline{0.0147}{\scriptsize$\pm$0.0016}
    & 0.0002{\scriptsize$\pm$0.0000}
    & 0.0109{\scriptsize$\pm$0.0068}
    & 0.0037{\scriptsize$\pm$0.0054}
    & 0.0322{\scriptsize$\pm$0.0547} \\

   & TimePrism
    & 0.0096{\scriptsize$\pm$0.0085}
    & 0.0250{\scriptsize$\pm$0.0164}
    & 0.0015{\scriptsize$\pm$0.0020}
    & 0.0294{\scriptsize$\pm$0.0255}
    & 0.0061{\scriptsize$\pm$0.0032}
    & 3.5007{\scriptsize$\pm$1.1821} \\

   & TACTiS-2
    & \textbf{0.0039}{\scriptsize$\pm$0.0000}
    & 0.0188{\scriptsize$\pm$0.0000}
    & \underline{0.0002}{\scriptsize$\pm$0.0000}
    & 0.0036{\scriptsize$\pm$0.0000}
    & 0.0005{\scriptsize$\pm$0.0000}
    & 0.0007{\scriptsize$\pm$0.0000} \\

   & TempFlow
    & 0.0057{\scriptsize$\pm$0.0001}
    & 0.0153{\scriptsize$\pm$0.0026}
    & 0.0003{\scriptsize$\pm$0.0000}
    & \underline{0.0032}{\scriptsize$\pm$0.0001}
    & \underline{0.0005}{\scriptsize$\pm$0.0000}
    & \underline{0.0005}{\scriptsize$\pm$0.0000} \\

   & Trans-TempFlow
    & 0.0060{\scriptsize$\pm$0.0002}
    & 0.0174{\scriptsize$\pm$0.0002}
    & 0.0003{\scriptsize$\pm$0.0000}
    & 0.0032{\scriptsize$\pm$0.0000}
    & 0.0005{\scriptsize$\pm$0.0000}
    & 0.0005{\scriptsize$\pm$0.0000} \\

   & TimeGrad
    & 0.0169{\scriptsize$\pm$0.0016}
    & 0.1753{\scriptsize$\pm$0.0826}
    & 0.0073{\scriptsize$\pm$0.0038}
    & 0.5454{\scriptsize$\pm$0.0505}
    & 0.1301{\scriptsize$\pm$0.0638}
    & 0.8028{\scriptsize$\pm$0.4729} \\

   & \textbf{PowerForge (Ours)}
    & \underline{0.0042}{\scriptsize$\pm$0.0019}
    & \textbf{0.0084}{\scriptsize$\pm$0.0025}
    & \textbf{0.0001}{\scriptsize$\pm$0.0001}
    & \textbf{0.0010}{\scriptsize$\pm$0.0002}
    & \textbf{0.0004}{\scriptsize$\pm$0.0000}
    & \textbf{0.0004}{\scriptsize$\pm$0.0000} \\
\midrule
  2869-bus & DeepAR
    & 0.1589{\scriptsize$\pm$0.0028}
    & 0.7711{\scriptsize$\pm$0.0301}
    & 0.0132{\scriptsize$\pm$0.0019}
    & 0.9145{\scriptsize$\pm$0.0083}
    & 0.8922{\scriptsize$\pm$0.0319}
    & 2.3529{\scriptsize$\pm$0.0919} \\

   & ETS
    & 0.0170{\scriptsize$\pm$0.0000}
    & 0.1041{\scriptsize$\pm$0.0000}
    & 0.0038{\scriptsize$\pm$0.0000}
    & 0.0640{\scriptsize$\pm$0.0005}
    & 0.0023{\scriptsize$\pm$0.0000}
    & 0.0106{\scriptsize$\pm$0.0000} \\

   & TimeMCL
    & 0.0235{\scriptsize$\pm$0.0016}
    & 0.0608{\scriptsize$\pm$0.0005}
    & 0.0039{\scriptsize$\pm$0.0002}
    & 0.0228{\scriptsize$\pm$0.0066}
    & 0.0130{\scriptsize$\pm$0.0055}
    & 0.1271{\scriptsize$\pm$0.0551} \\

   & TimePrism
    & 0.0222{\scriptsize$\pm$0.0018}
    & \underline{0.0593}{\scriptsize$\pm$0.0026}
    & 0.0036{\scriptsize$\pm$0.0003}
    & 0.0862{\scriptsize$\pm$0.0457}
    & 0.0018{\scriptsize$\pm$0.0011}
    & 2.5224{\scriptsize$\pm$1.3747} \\

   & TACTiS-2
    & \underline{0.0154}{\scriptsize$\pm$0.0001}
    & 0.0844{\scriptsize$\pm$0.0038}
    & \underline{0.0036}{\scriptsize$\pm$0.0000}
    & 0.0069{\scriptsize$\pm$0.0004}
    & \underline{0.0002}{\scriptsize$\pm$0.0000}
    & 0.0004{\scriptsize$\pm$0.0000} \\

   & TempFlow
    & 0.0175{\scriptsize$\pm$0.0019}
    & 0.0679{\scriptsize$\pm$0.0040}
    & 0.0037{\scriptsize$\pm$0.0001}
    & \underline{0.0062}{\scriptsize$\pm$0.0003}
    & 0.0003{\scriptsize$\pm$0.0000}
    & \underline{0.0003}{\scriptsize$\pm$0.0000} \\

   & Trans-TempFlow
    & 0.0249{\scriptsize$\pm$0.0069}
    & 0.0940{\scriptsize$\pm$0.0238}
    & 0.0046{\scriptsize$\pm$0.0007}
    & 0.3051{\scriptsize$\pm$0.2633}
    & 0.0105{\scriptsize$\pm$0.0095}
    & 0.0263{\scriptsize$\pm$0.0228} \\

   & TimeGrad
    & 0.0269{\scriptsize$\pm$0.0012}
    & 0.1707{\scriptsize$\pm$0.0255}
    & 0.0123{\scriptsize$\pm$0.0013}
    & 0.1201{\scriptsize$\pm$0.0115}
    & 0.0162{\scriptsize$\pm$0.0139}
    & 0.1361{\scriptsize$\pm$0.1135} \\

   & \textbf{PowerForge (Ours)}
    & \textbf{0.0112}{\scriptsize$\pm$0.0004}
    & \textbf{0.0344}{\scriptsize$\pm$0.0011}
    & \textbf{0.0012}{\scriptsize$\pm$0.0001}
    & \textbf{0.0025}{\scriptsize$\pm$0.0000}
    & \textbf{0.0002}{\scriptsize$\pm$0.0000}
    & \textbf{0.0002}{\scriptsize$\pm$0.0000} \\
\midrule
  3120-bus & DeepAR
    & 0.1408{\scriptsize$\pm$0.0148}
    & 0.7666{\scriptsize$\pm$0.0842}
    & 0.0093{\scriptsize$\pm$0.0017}
    & 0.7883{\scriptsize$\pm$0.0160}
    & 0.6322{\scriptsize$\pm$0.2054}
    & 1.6903{\scriptsize$\pm$0.5156} \\

   & ETS
    & 0.0180{\scriptsize$\pm$0.0001}
    & 0.1954{\scriptsize$\pm$0.0000}
    & 0.0075{\scriptsize$\pm$0.0000}
    & 0.0987{\scriptsize$\pm$0.0001}
    & 0.0069{\scriptsize$\pm$0.0000}
    & 0.0225{\scriptsize$\pm$0.0000} \\

   & TimeMCL
    & 0.0092{\scriptsize$\pm$0.0052}
    & \underline{0.0162}{\scriptsize$\pm$0.0044}
    & 0.0004{\scriptsize$\pm$0.0004}
    & 0.1100{\scriptsize$\pm$0.1071}
    & 0.0252{\scriptsize$\pm$0.0242}
    & 0.2240{\scriptsize$\pm$0.2385} \\

   & TimePrism
    & 0.0083{\scriptsize$\pm$0.0018}
    & 0.0213{\scriptsize$\pm$0.0119}
    & 0.0013{\scriptsize$\pm$0.0006}
    & 0.0579{\scriptsize$\pm$0.0383}
    & 0.0043{\scriptsize$\pm$0.0017}
    & 5.4230{\scriptsize$\pm$3.2742} \\

   & TACTiS-2
    & \underline{0.0043}{\scriptsize$\pm$0.0000}
    & 0.0297{\scriptsize$\pm$0.0001}
    & \underline{0.0002}{\scriptsize$\pm$0.0000}
    & 0.0428{\scriptsize$\pm$0.0006}
    & 0.0033{\scriptsize$\pm$0.0001}
    & 0.0070{\scriptsize$\pm$0.0000} \\

   & TempFlow
    & 0.0056{\scriptsize$\pm$0.0001}
    & 0.0178{\scriptsize$\pm$0.0001}
    & 0.0003{\scriptsize$\pm$0.0000}
    & 0.0188{\scriptsize$\pm$0.0003}
    & 0.0033{\scriptsize$\pm$0.0000}
    & 0.0035{\scriptsize$\pm$0.0000} \\

   & Trans-TempFlow
    & 0.0054{\scriptsize$\pm$0.0002}
    & 0.0175{\scriptsize$\pm$0.0006}
    & 0.0003{\scriptsize$\pm$0.0000}
    & \underline{0.0184}{\scriptsize$\pm$0.0004}
    & \underline{0.0032}{\scriptsize$\pm$0.0000}
    & \underline{0.0034}{\scriptsize$\pm$0.0000} \\

   & TimeGrad
    & 0.0123{\scriptsize$\pm$0.0069}
    & 0.1295{\scriptsize$\pm$0.0503}
    & 0.0035{\scriptsize$\pm$0.0039}
    & 0.4234{\scriptsize$\pm$0.1359}
    & 0.0887{\scriptsize$\pm$0.0508}
    & 0.5695{\scriptsize$\pm$0.2468} \\

   & \textbf{PowerForge (Ours)}
    & \textbf{0.0038}{\scriptsize$\pm$0.0011}
    & \textbf{0.0076}{\scriptsize$\pm$0.0010}
    & \textbf{0.0001}{\scriptsize$\pm$0.0000}
    & \textbf{0.0068}{\scriptsize$\pm$0.0000}
    & \textbf{0.0025}{\scriptsize$\pm$0.0000}
    & \textbf{0.0025}{\scriptsize$\pm$0.0000} \\
\bottomrule
\end{tabular}%
}
\end{table*}

\subsubsection{Optimisation and Checkpointing}
\label{app:optim}

PowerForge is trained with Adam at learning rate $10^{-3}$ and weight decay $10^{-6}$, for up to $200$ epochs per network. Validation loss is computed on the held-out validation windows after each epoch, and the checkpoint with the lowest validation loss is used for evaluation. The learning rate is halved by \texttt{ReduceLROnPlateau} when validation loss plateaus.

\subsubsection{Baseline Configurations}
\label{app:baselines}

Deep baselines (DeepAR, TempFlow, Transformer-TempFlow, TimeGrad, TACTiS-2, TimeMCL, TimePrism) follow their GluonTS \citep{alexandrov2020gluonts} or original-release reference implementations.  We use each baseline's default optimiser, learning rate schedule, and training epoch budget as specified in its reference codebase, with TACTiS-2 trained on its first stage only due to compute constraints at our scale. We apply the same train, validation, and test splits as PowerForge (Appendix~\ref{app:splits}). ETS is fitted per channel by maximum likelihood on CPU.

\section{Full Results on the Five Main Grids}
\label{app:full_table}

Table~\ref{tab:powerphase_five_grids} reports six metrics on every model
across the five main PowerPhase grids, extending the
three-grid scoreboard in Table~\ref{tab:main}. 

The pattern from the representative grids holds across the full set. PowerForge attains the best score on every metric on PEGASE\,1354, PEGASE\,2869, and Polish\,3120, the best Distortion, Safety\_mBrier, and $\mathrm{CVaR}_{0.1}$ on Polish\,2383 with CRPS within seed variance of the strongest baseline, and ties with TACTiS-2 on Safety\_mBrier and $\mathrm{CVaR}_{0.1}$ at ACTIVSg\,500 where both achieve zero violations. The two intermediate grids not shown in the main
text (PEGASE\,1354, PEGASE\,2869) follow the same ordering.

\section{Evaluation at 36,964 Channels: PEGASE 9241}
\label{app:9241}

This appendix evaluates PowerForge at the upper end of the PowerPhase scale, on PEGASE 9241 (36{,}964 channels). The five-grid evaluation in Section~\ref{exp1} characterises PowerForge against the full eight-baseline panel. Here we extend the comparison at maximum scale to four baselines spanning three families: TimeMCL and TimePrism (scenario-based), Transformer-TempFlow (conditioned normalising flow), and ETS (statistical reference). DeepAR, TimeGrad, TACTiS-2, and TempFlow are not included in this comparison. TempFlow's behavior closely tracks Trans-TempFlow on the five main grids. PowerForge, TimePrism, TimeMCL, ETS and Transformer-TempFlow are evaluated with three random seeds $\{22, 42, 3142\}$.

\begin{table}[h]
\caption{Probabilistic forecasting results on PEGASE 9241 (36{,}964 channels).
CRPS and Distortion measure \emph{fidelity}; Safety\_mBrier and $\text{CVaR}_{0.1}$ measure \emph{safety}.
Values are mean$\pm$std over three seeds.
\textbf{Bold}: best; \underline{underline}: second best.}
\label{tab:9241}
\centering
\small
\begin{tabular}{@{}l cc cc@{}}
\toprule
 & \multicolumn{2}{c}{\emph{Fidelity}\,$\downarrow$}
 & \multicolumn{2}{c}{\emph{Safety}\,$\downarrow$} \\
\cmidrule(lr){2-3} \cmidrule(lr){4-5}
\textbf{Model}
  & CRPS & Distortion
  & Safety\_mBrier & $\text{CVaR}_{0.1}$ \\
\midrule
ETS                       & 0.0413\,$\pm$\,0.0001 & 0.4515\,$\pm$\,0.0009 & 0.1361\,$\pm$\,0.0001 & 0.0395\,$\pm$\,0.0001 \\
TimeMCL                   & 0.0349\,$\pm$\,0.0129 & 0.0629\,$\pm$\,0.0030 & 0.2326\,$\pm$\,0.3072 & 0.4726\,$\pm$\,0.4843 \\
TimePrism                 & \underline{0.0112}\,$\pm$\,0.0029 & \underline{0.0427}\,$\pm$\,0.0046 & 0.0506\,$\pm$\,0.0393 & 1.5350\,$\pm$\,0.3220 \\
Trans-TempFlow & 0.0224\,$\pm$\,0.0019 & 0.0748\,$\pm$\,0.0039 & \underline{0.0087}\,$\pm$\,0.0013 & \underline{0.0029}\,$\pm$\,0.0001 \\
\midrule
\textbf{PowerForge (Ours)}
  & \textbf{0.0077}\,$\pm$\,0.0008
  & \textbf{0.0185}\,$\pm$\,0.0024
  & \textbf{0.0021}\,$\pm$\,0.0000
  & \textbf{0.0023}\,$\pm$\,0.0000 \\
\bottomrule
\end{tabular}
\end{table}

PowerForge attains the best score on every metric. Two observations transfer the pattern of Section~\ref{exp1} to the maximum-scale setting. First, the safety--fidelity trade-off remains visible: TimePrism is competitive on CRPS (0.0112) but its $\mathrm{CVaR}_{0.1}$ degrades to 1.535, indicating that its hypothesis spread covers infeasible voltage states with substantial probability. Trans-TempFlow follows the inverse pattern observed at 3120-bus, with strong safety scores but weaker fidelity than PowerForge. Second, ETS scales to $36{,}964$ channels through per-channel CPU fitting and provides a non-trivial statistical reference, but its CRPS and Distortion sit an order of magnitude above the deep baselines, consistent with the pattern observed on the five main grids.

\section{Additional Safety Analyses}
\label{app:additional_safety}

\subsection{Additional Qualitative Comparison: PEGASE 1354}
\label{app:additional_qualitative}
\begin{figure}[t]
    \centering
    \includegraphics[width=\textwidth]{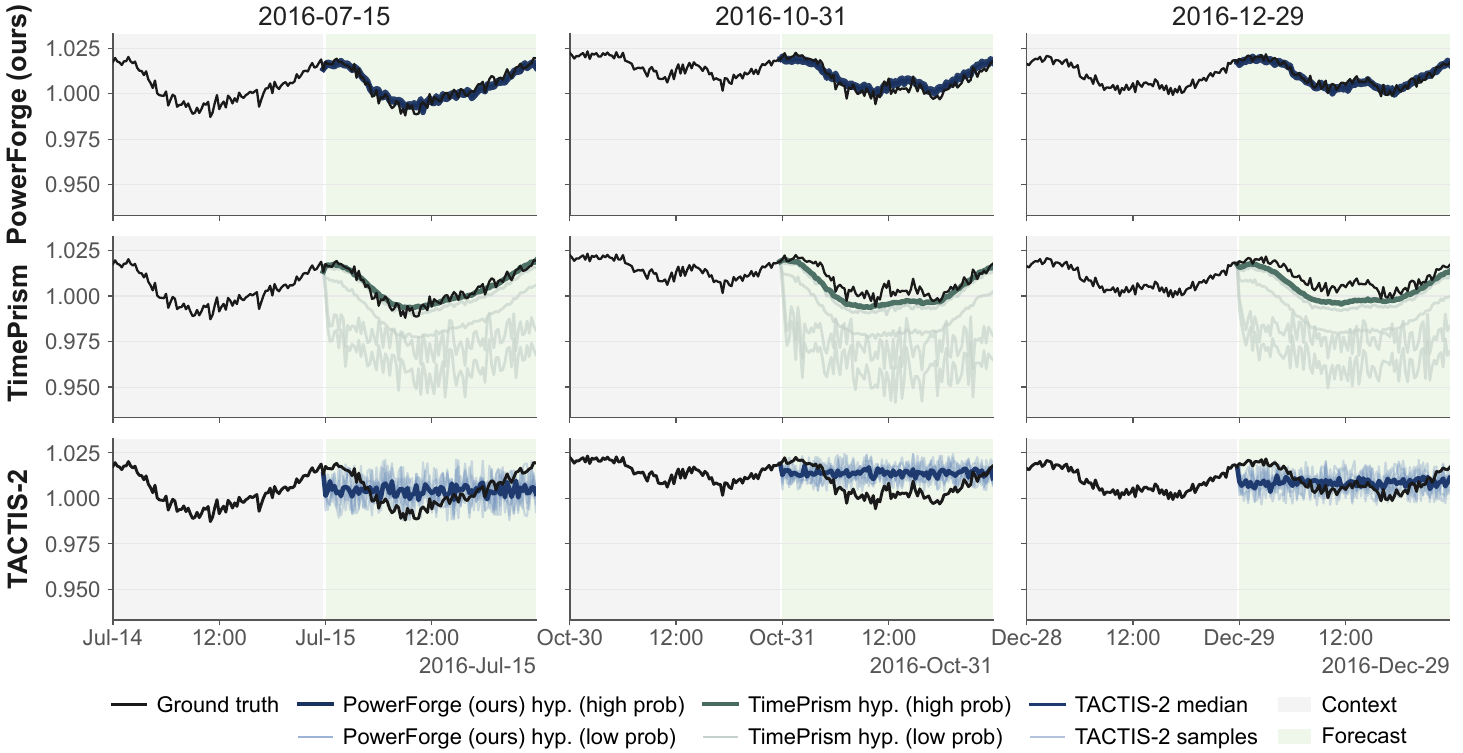}
    \caption{Voltage forecasts on PEGASE\,1354 ($5{,}416$ channels) for a single bus across three test windows. The three-row structure mirrors Figure~\ref{fig:qualitative}. This bus operates in the nominal voltage band with a morning diurnal trough, in contrast to the boundary-crossing evening-trough bus in Figure~\ref{fig:qualitative}. TACTiS-2's median tracks the morning dip across all three windows while samples retain high-frequency jitter, and the qualitative ordering across forecasters is preserved relative to the Polish 2383 case.}
    \label{fig:qualitative_1354}
\end{figure}

Figure~\ref{fig:qualitative_1354} reports the same three-row qualitative comparison on PEGASE\,1354 ($5{,}416$ channels). PowerForge produces a compact hypothesis set that tracks the diurnal trajectory, TimePrism recovers the shape with a wider spread, and TACTiS-2's median tracks the morning trough on this nominal-band bus while individual samples retain high-frequency jitter. The qualitative ordering across forecasters is preserved relative to the Polish 2383 case in the main text, although the per-step sample jitter of TACTiS-2 is more salient than its low-frequency flattening on this particular bus.

\subsection{Bus-level aggregate distribution on Polish 2383}
\label{app:additional_safety:2383}

\begin{figure}[t]
    \centering
    \includegraphics[width=\textwidth]{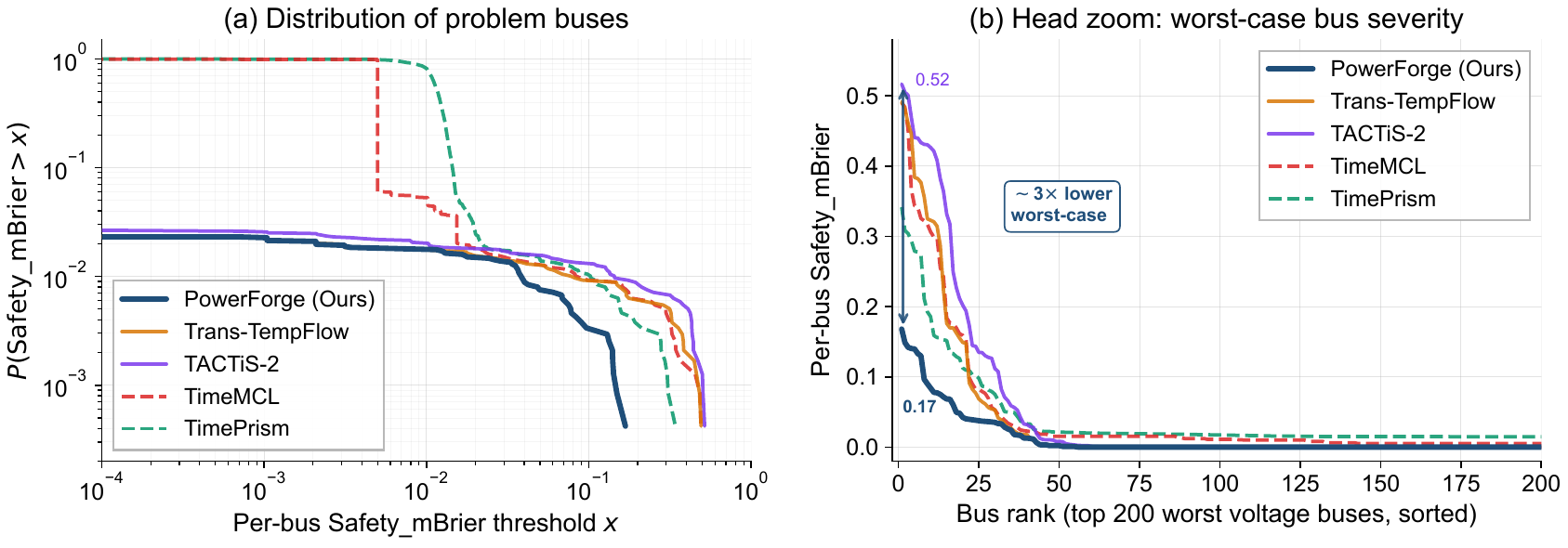}
    \caption{Bus-level aggregate voltage safety on Polish\,2383 ($9{,}532$ channels), computed as per-bus Safety$\_$mBrier averaged across all forecast scenarios, time steps, and ten test windows. \textbf{(a)} Survival function of per-bus Safety$\_$mBrier. PowerForge, Trans-TempFlow, and TACTiS-2 (solid) keep $\sim$97\% of 2{,}383 buses below any non-trivial threshold, while TimeMCL and TimePrism (dashed) place violation probability on nearly all buses. \textbf{(b)} Top-200 worst buses, sorted by severity. PowerForge top-1 bus 0.17 vs 0.52 (TACTiS-2), an $\sim$$3\times$ improvement on the hardest bus. ETS, DeepAR, and TimeGrad are omitted (full results in Table~\ref{tab:powerphase_five_grids}).}
    \label{fig:aggregate-safety-2383}
\end{figure}

Section~\ref{app:additional_qualitative} extends the channel-level analysis of Section~\ref{sec:qualitative} to PEGASE 1354. Here we complement the channel-level view with a network-level distributional analysis on Polish\,2383, the grid used for the main qualitative figure. The per-bus Safety$\_$mBrier (Figure~\ref{fig:aggregate-safety-2383}) measures how often and how confidently each model places probability mass outside the $[0.95, 1.05]$ p.u.\ voltage band at each of the 2{,}383 voltage buses.

\paragraph{Panel (a): distribution of problem buses.} The horizontal axis is a per-bus Safety$\_$mBrier threshold $x$ on log scale, and the vertical axis is the fraction of buses whose per-bus value exceeds $x$. Two regimes emerge. PowerForge, Trans-TempFlow, and TACTiS-2 (solid) sit near $y \approx 0.025$ even at very small $x$, meaning roughly 97\% of buses are forecast cleanly and violations concentrate on a small problem set. TimeMCL and TimePrism (dashed) instead remain near $y = 1.0$ until $x \approx 10^{-2}$, indicating pervasive low-magnitude violations across almost all buses. This separation matches the safety-fidelity trade-off in Table~\ref{tab:powerphase_five_grids}.

\paragraph{Panel (b): worst-case bus severity.} The horizontal axis is the rank of the worst 200 buses per model, sorted by descending per-bus Safety$\_$mBrier. Among the three sparse-violation methods, PowerForge has top-1 bus severity 0.17 versus 0.49 (Trans-TempFlow) and 0.52 (TACTiS-2). PowerForge's curve also stays consistently below the other four models across the full top-200 head.

\paragraph{Network-determined hard buses.} The hard-bus set is largely shared across the three sparse-violation models. The top-50 problem buses overlap 100\% between PowerForge and Trans-TempFlow, and 96\% between PowerForge and TACTiS-2. Per-bus Safety$\_$mBrier rankings across all 2{,}383 buses are also strongly correlated, with Spearman $\rho > 0.999$ between PowerForge and Trans-TempFlow and $\rho = 0.928$ between PowerForge and TACTiS-2. The hard-bus set is therefore largely determined by the forecasting task on this network rather than by individual model choices. PowerForge does not eliminate the hard-bus phenomenon, but reduces worst-case severity on the same network-determined set.

\section{Full Ablation Results on PEGASE 1354}
\label{app:ablation_1354}

\begin{table}[h]
\centering
\small
\setlength{\tabcolsep}{4pt}
\caption{Ablation on PEGASE~1354 (5{,}416 channels), single seed. $\Delta$ is the relative CRPS change against the full configuration. Variants are ordered by descending CRPS impact. Lower is better for all metrics.}
\label{tab:ablation_1354}
\begin{tabular}{lrrrrrr}
\toprule
Variant & CRPS $\downarrow$ & $\Delta$ & Distortion $\downarrow$ & Safety\_mBrier $\downarrow$ & NECV $\downarrow$ & CVaR$_{0.1}$ $\downarrow$ \\
\midrule
Full PowerForge              & 0.00433 & --      & 0.0110 & 0.00301 & 7.41e-4 & 7.52e-4 \\
w/o anchor-delta             & 0.00797 & +84\%  & 0.0229 & 0.03042 & 8.09e-4 & 1.90e-3 \\
w/o quantile regularisation  & 0.00594 & +37\%  & 0.0128 & 0.00310 & 7.41e-4 & 7.54e-4 \\
w/o ordered-quantile (WTA)   & 0.00551 & +27\%  & 0.0155 & 0.00448 & 7.46e-4 & 8.71e-4 \\
w/o physics regularisation   & 0.00550 & +27\%  & 0.0120 & 0.00308 & 7.39e-4 & 7.52e-4 \\
w/o cross-type causal bridge & 0.00530 & +22\%  & 0.0120 & 0.00307 & 7.40e-4 & 7.53e-4 \\
w/o low-rank global mixer    & 0.00450 & +4\%   & 0.0154 & 0.00303 & 7.36e-4 & 7.42e-4 \\
\bottomrule
\end{tabular}
\end{table}

Table~\ref{tab:ablation_1354} extends the PEGASE~1354 ablation in Section~\ref{sec:ablation} to the full PowerPhase metric suite (CRPS, Distortion, Safety\_mBrier, NECV, CVaR$_{0.1}$) under a single random seed. The safety metrics identify anchor-delta as the dominant component for voltage-band feasibility, while the remaining variants leave Safety\_mBrier, NECV, and CVaR$_{0.1}$ close to the full-model baseline.









\end{document}